\documentclass[a4paper]{article}

\usepackage{coolArticle}
\usepackage{coolMath}
\usepackage{xcolor}
\usepackage{authblk}
\usepackage{breakcites}
\usepackage{subcaption}
\usepackage{slashbox}

\graphicspath{{./img/}}
\definecolor{lightblue}{rgb}{0.88,0.9,1}

\title{\vhrulefill{5pt} \\ \vspace{5pt} \textbf{Rover Descent: Learning to optimize by learning to navigate on prototypical loss surfaces} \\ \vspace{5pt}\vhrulefill{2pt}}

\author[1,2]{Louis Faury \thanks{\texttt{l.faury@criteo.com}}}
\author[1]{Flavian Vasile \thanks{\texttt{f.vasile@criteo.com}}}
\affil[1]{Criteo Research, France}
\affil[2]{Ecole Polytechnique Federale de Lausanne, Switzerland}
\date{}
\begin{document}

\maketitle

\renewcommand{\abstractname}{}
\begin{abstract}
Learning to optimize - the idea that we can learn from data algorithms that optimize a numerical criterion - has recently been at the heart of a growing number of research efforts. One of the most challenging issues within this approach is to learn a policy that is able to optimize over classes of functions that are different from the classes that the policy was trained on. We propose a novel way of framing learning to optimize as a problem of learning a good navigation policy on a partially observable loss surface. To this end, we develop \emph{Rover Descent}, a solution that allows us to learn a broad optimization policy from training only on a small set of \emph{prototypical two-dimensional surfaces} that encompasses classically hard cases such as valleys, plateaus, cliffs and saddles and by using strictly zeroth-order information. We show that, without having access to gradient or curvature information, we achieve fast convergence on optimization problems not presented at training time, such as the Rosenbrock function and other two dimensional hard functions.  We extend our framework to optimize over high dimensional functions and show good preliminary results.
\end{abstract}

\section{Introduction}
{
	Finding the minimizer $\theta^*$ of a function $f$ over some domain $\Omega$ is a recurrent problem in a large variety of engineering and scientific tasks. Instances of this problem appear in machine learning, optimal control, inventory management, portfolio optimization, and many other applications. This great diversity of problems has led over the years to the development of a large body of optimization algorithms, from very general to problem-specific ones. 
	
	Recently, the advent of deep learning led to the creation of several methods targeting high-dimensional, non-convex problems (the most famous ones being momentum \cite{nesterov1983method}, Adadelta \cite{zeiler2012adadelta} and Adam \cite{kingma2014adam}), now used as black-box algorithms by a majority of practitioners. Other attempts in this field use some additional problem-specific structure, like the work by \cite{martens2010deep} that leverages fast multiplication by the Hessian to yield better performing optimization policies, though computationally demanding. A common point to all these algorithms is that they leverage human-based understanding of loss surfaces, and usually require tuning hyper-parameters to achieve state-of-the-art performance. This tuning process can sometimes reveal mysterious behavior of the handled optimizers, making it reserved to human experts or the subject of a long and tedious search. Also, the process results in a static optimizer which excels at the specific task, but is likely to perform poorly on others. 
	
	If the limitations of hand-designed algorithms come from poor human understanding of high-dimensional loss landscapes, it is natural to ask what machine learning can do for the design of optimization algorithms. Recently, \cite{andrychowicz2016learning} and \cite{li2016learning} both introduced two frameworks for learning optimization algorithms. While the former proposes to learn task-specific optimizers, the latter aims to produce task-independent optimization policies. While in the most general case this is bound to fail - as suggested by the No Free Lunch theorem for combinatorial optimization \cite{wolpert1997no} - we also believe that data driven techniques can be robust on a great variety of problems.
	
	Most optimization algorithms can be framed, for a given objective function $f$ and a current iterate $\theta_i$, as the problem of finding an appropriate update $\Delta \theta_i$. This update can for instance depend on past gradient information, rescaled gradient using curvature information or many other features. In a general manner, we can write $\Delta \theta_i = \phi( \theta_i, h(f,\theta_{i-1},\hdots,\theta_0),\xi)$ where $h(\cdot)$ denotes the set of features accumulated during the optimization procedure, and $\xi$ denotes the optimization hyper-parameters. 
		
   	In our approach, we aim to bypass computing gradient and curvature information and learn the optimization features directly from data. This should allows us to obtain local state descriptors that can outperform classical features in terms of generalization on unseen loss functions and input data distributions. In this vein, we draw an analogy between learning an optimization algorithm and learning a navigation policy while having access to raw local observations of the landscape, which is also the inspiration for the name of our method, \emph{Rover Descent}. 
	Our algorithm contains three chained predictors that compute the angle of the move, the magnitude of the move (e.g. learning rate) and the resolution of the grid of the zeroth-order samples at the landing point. We train our navigation agent on hard \emph{prototypical 2D surfaces} in order to make sure we develop feature detectors and subsequent policies that will be able to lead to good decisions in difficult areas of the loss function. We pose both the learning rate and resolution predictor as reinforcement learning problems and introduce a reward-shaping formula that allows us to learn from functions with different magnitude and from multiple proto-families. In our experiments this was a crucial factor in being able to generalize on many different types of evaluation functions.
   	
	We show that this setup leads to very good convergence speeds both in two and higher dimensions, on evaluation functions that are not presented at training time. For a zeroth-order optimization algorithm, the convergence performance is surprisingly good, leading to results comparative to or better than the task-specific optimizer (\emph{e.g} the best one out of set of specifically tuned first and second order optimizers).
	
	In conclusion, we believe that our main contributions are the following: framing the learning to optimize problem as a navigation task,  proposing a zeroth-order information-based learning architecture, coupled with a proper training procedure on prototypical two-dimensional surfaces and a reward shaping formula and showing experimentally that it can match/outperform first and second order techniques on meta-generalization tasks.
	
	The rest of the paper is organized as follows. We first give a brief summary of past and recent related work in the field of learning to learn and learning to optimize, and position our approach with respect to existing work in Section \ref{sec::related}. We then develop in \ref{sec::twod} our approach in the two-dimensional case, before extending it to a higher dimensional setting in \ref{sec::highd}. We present experimental results in Section \ref{sec::experiments} that show the validity of our approach in a variety of setups. We finally develop potential ideas for future work in Section \ref{sec::conclusion}. 
}


\section{Related work}
{	\label{sec::related}
	\paragraph{} The field of optimization has been studied for many years and for a great diversity of problems. Providing a complete review of the subject would be out of the scope of this paper, and therefore we provide only a short reminder on the different approaches of the domain.  
	
	Some simple settings (convexity, $L$-smoothness, ..) have been intensively exploited to devise a large number of optimizers, derive upper-bound convergence rates (\cite{nesterov2013introductory} and \cite{nemirovskii1983problem}) and even some information theoretical complexity lower bounds for black-box optimizers \cite{agarwal2009information}. More recently, motivated by the growing interest in deep learning, a lot of research efforts were also invested in devising smart, adaptative optimizers for complicated, very high dimensional objectives. 
	
	Part of the diversity of existing optimizers is explained by the different type of oracles (possibly noisy, second, first, zeroth order evaluation oracles or even comparison oracle) available for a given problem. The case of noisy first order oracles has been widely adopted in the machine learning community and led to many innovations (a detailed survey can be found in \cite{bottou2016optimization}). Noisy zeroth order oracles also received a lot of attention from the bandit, the Bayesian optimization and the evolutionary optimization communities (one of the most successful method being Covariance Matrix Adaptation Evolutionary Strategy \cite{hansen2016cma}), and have also seen a few heuristics approaches (the Nelder-Meald heuristic \cite{nelder1965simplex} being one of them). 
	
	\paragraph{} Learning to learn (or meta-learning) is not a recent idea. \cite{schmidhuber1987} thought of a Recurrent Neural Network (RNN) able to modify its own weights, building a fully differentiable system allowing the training to be learned by gradient descent. \cite{hochl2l} proposed to discover optimizers by gradient descent, optimizing RNNs modeling the optimization sequence with a learning signal emerging from backpropagation on a first network. 
	
	Recently, some meta-learning tentatives have shown great progress in different optimization fields. Various attempt tried to dynamically adapt the hyper-parameters of hand-designed algorithms, like \cite{daniel2016learning} or \cite{hansen2016using}. Using gradient statistics as an input for a recurrent neural net, \cite{li2017learning} were able to reinforcement learn a policy effective for training deep neural networks. In \cite{andrychowicz2016learning}, the authors show that when leveraging first-order information one could learn by gradient-descent optimizers that outperforms current state-of-the-art of existing problems - however only when the meta-train dataset is made of the same class of problem. When confronted to a different class of functions, the meta-learner is unable to infer efficient optimization moves. With the same idea of using gradient-descent for training the optimizer, \cite{chen2017learning} use zeroth order information in order to learn an optimizer for the Bayesian optimization setting.  

	However, one could think that by showing enough examples to a meta-learner (namely made up of instances where traditional optimizers reach their limits), and adapting its structure to cover a large number of classes of functions, it could adapt to unknown loss landscapes. This idea was exploited by \cite{wil2l}, who manage to learn optimizers that generalizes to completely unseen data, while still being able to scale up to high-dimensional problems. Their process namely involves training by gradient descent hierarchical RNNs and showing it a great variety of examples. However, their optimizer's structure remains quite complicated, and doesn't provide human-level understanding of the features leveraged by the meta-learner. We believe that a more intelligible architecture could enable us to understand better what the network is learning, while still being effective on a large class of functions, when trained on a selected number of meta-examples. 
}

\section{Our approach}

\subsection{Intuition}
	
	In this paper, we propose framing the problem of learning to optimize as a problem of navigation on the partially observable error surface. The error surface is defined by the values of the loss function taken over the range of its inputs. In this framework, the optimizer is an agent that starting from the initial point, attempts to reach the lowest point on the surface with the smallest number of actions (where an action is a move to an arbitrary point on the landscape), while observing only a set of points sampled from the loss surface. Our goal is to learn the navigation policy that maps the current state of the agent to a move on the surface. 
	
\begin{figure}[h!]
				\begin{center}
					\includegraphics[width=0.7\linewidth]{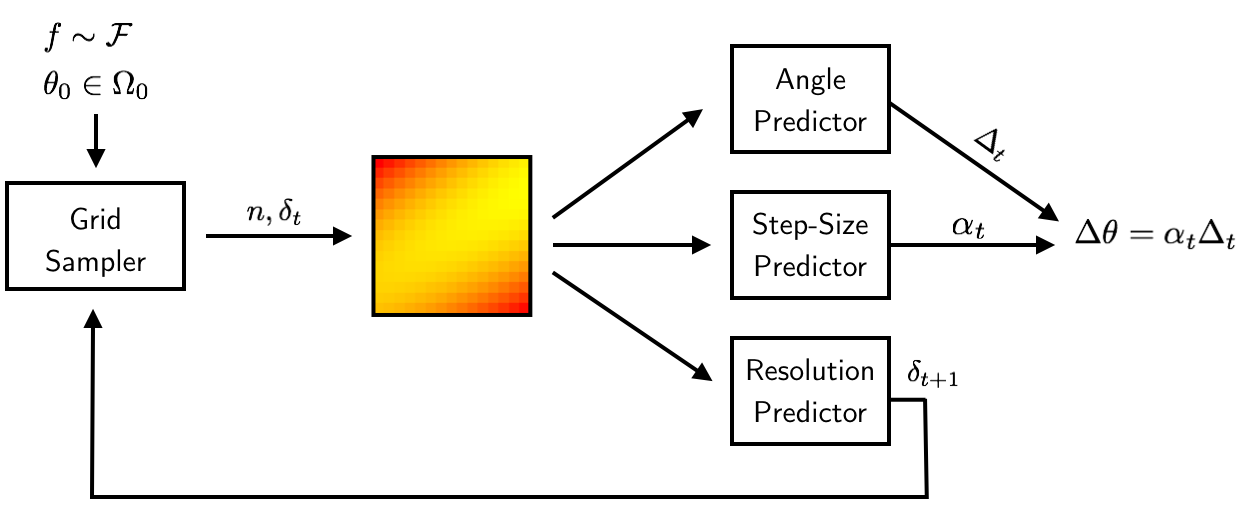}
					\caption{Decomposition of the optimization step in three independent modules: angle prediction, step-size prediction and resolution prediction.}
					\label{fig::archi}
				\end{center}
			\end{figure}

	To this end, we decide to divide the architecture of our agent in three sequential modules: the normalized update direction predictor $\Delta$, that predicts the angle of the update, the learning rate predictor that predicts the magnitude of the update $\alpha$, and the resolution predictor, that predicts the scale $\delta$ of the observation set at the landing point. This choice is motivated by our intuition that these steps can be approached in a hierarchical way (first choose a direction, then a step size accordingly for instance) and therefore might involve different training methods and procedures. Furthermore, each of the modules can act as a correction factor on the other two modules. For example, if the update angle is not correct, the learning rate module can compensate by making the move very small and the resolution module can zoom out/in to make the next angle prediction task easier. Figure \ref{fig::archi} sums up the architecture we just devised in the two-dimensional case.
	
			
	
\subsection{Architecture: the two-dimensional case}
\label{sec::twod}
In the following subsection, we consider the simple case $d=2$ to develop our experimental set-up. A generalization for higher dimensions can be found in Section \ref{sec::highd}. 
		
\subsubsection{Choosing the prototypical landscapes and the input representation}
	
\begin{figure}[h!]
				\begin{center}
					\includegraphics[width=0.9\linewidth]{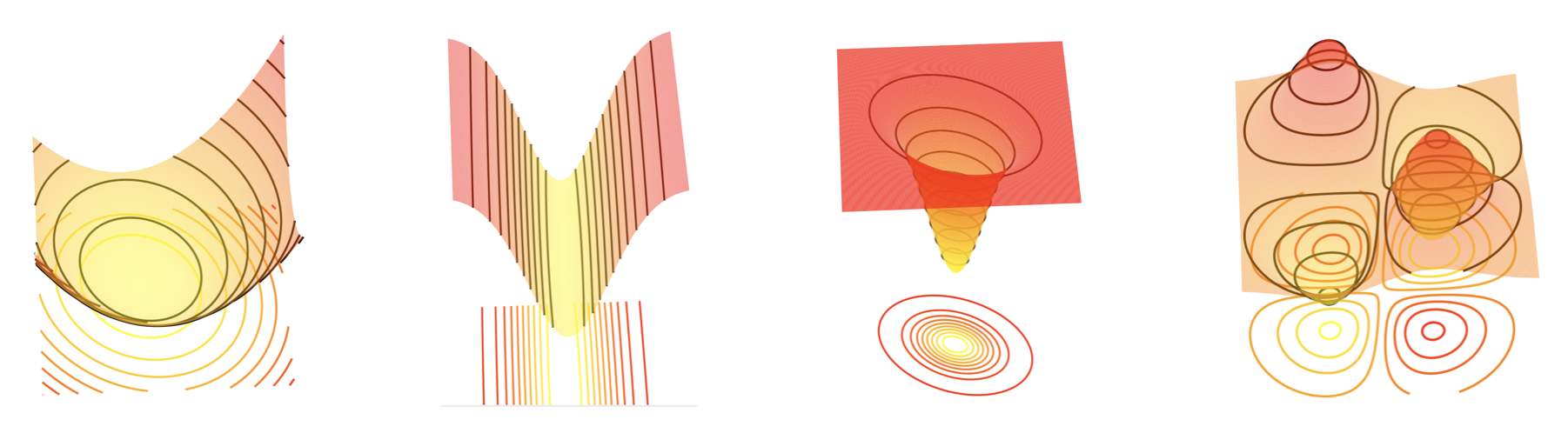}
					\caption{Instances of $\mathcal{F}_{train}$. In order: quadratic bowl, valley, (plateau+cliff), saddle. Best viewed in color. }
					\label{fig::sample_from_metatrain}
				\end{center}
			\end{figure}

	\paragraph{Prototypical landscapes} Because our end goal is to be able to optimize complex loss landscapes, we are interested in selecting a small but sufficiently large set of prototypical landscapes as our meta-training set $\mathcal{F}_{train}$. More precisely, we decide to target surface degeneracies that are common when learning the weights of deep neural networks. These namely include \emph{valleys}, \emph{plateaus}, but also \emph{cliffs} (\cite{bengio1994learning}) and \emph{saddles} (\cite{dauphin2014identifying}). We also consider it useful to add \emph{quadratic bowls} to that list, to provide simpler and saner landscapes. Figure \ref{fig::sample_from_metatrain} provides a visualisation of each of these landscapes as generated by our meta-training algorithm, for which details can be found in Appendix \ref{sec::landscape_detail}. Interestingly enough, all of these landscapes were listed in \cite{schaul2013unit}, which provides a collection of unit tests for optimization. In the line of this work, we estimate that learning a optimizer over such landscapes can result in a robust algorithm. Also, because it is frequent in real world applications to only have access to noisy samples of the function we wish to optimize, our framework should therefore provide noisy versions of the landscapes described hereinbefore. 
     		
\begin{figure}[h!]
			\begin{center}
				\includegraphics[width=0.7\linewidth]{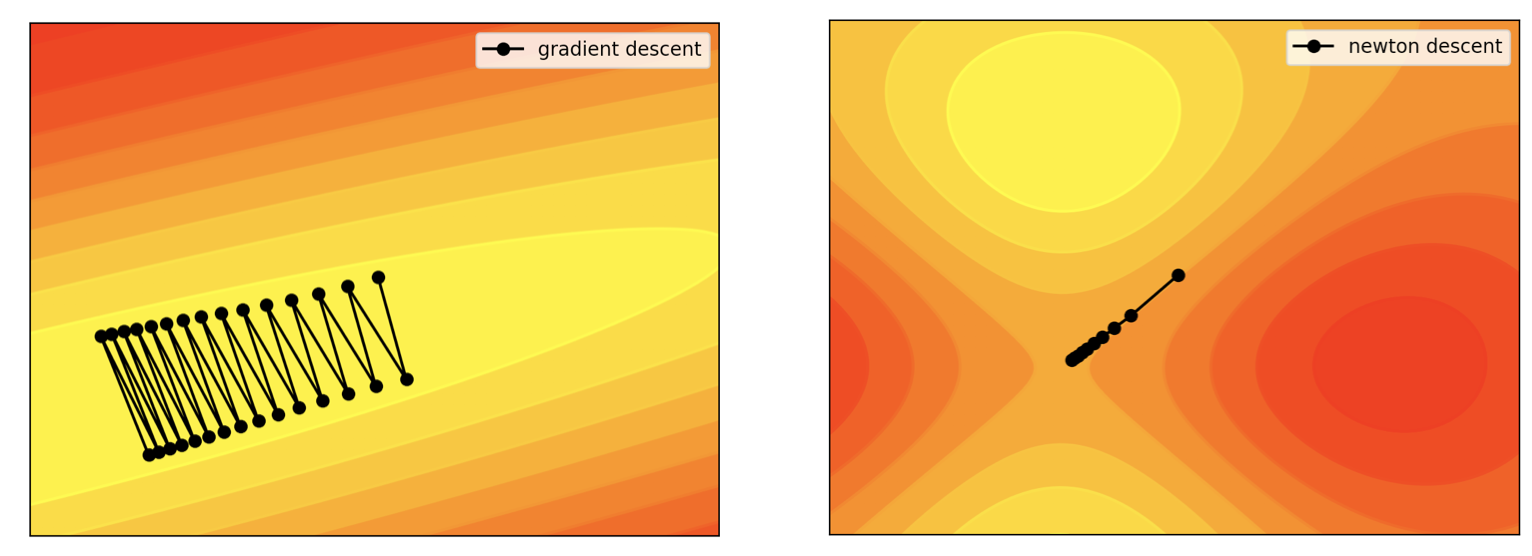}
				\caption{\emph{Left}: gradient descent has a shattering behavior in narrow valleys. \emph{Right}: saddle points are attractors for Newton descent.}
				\label{fig::nemesis}
			\end{center}
		\end{figure}

	\paragraph{Input design} Let us consider two classical optimizers: gradient-descent and Newton descent (see \cite{nocedal2006numerical} for complete details). While gradient descend can escape non-strict saddle points but shows shattering behavior insides valleys (see Figure \ref{fig::nemesis}), second-order methods can leverage curvature to make quick progress inside valleys. However, saddles are attraction points for such methods. 
	
	To get the best of both worlds, we want our local descriptor to be able to represent both first and second order information. Finite difference provides an easy way to approximate them from zeroth order sample of a function. However, the precision of finite difference can  be severely impacted by noisy oracles, although this can be alleviated by a pre-filtering of the function samples (like low-pass filtering for removing white noise). 
			
		Let $f$ the loss landscape we are optimizing, mapping $\Omega_f \subset \mathbb{R}^d$ into $\mathbb{R}$, and $\theta \in\Omega_f$. A natural way to describe the surroundings of $\theta$ is to sample a grid centered on $\theta$. Given a budget of $n^2$ samples and a resolution $\delta$, we note $s_\delta^n(f,\theta)$ the resulting two-dimensional grid.
		\begin{equation}
				s_\delta^n(f,\theta) \triangleq \left(f \left( \begin{pmatrix} \theta_1 \\ \theta_2 \end{pmatrix} - \delta\cdot\begin{pmatrix} i - n/2 \\ j-n/2\end{pmatrix}\right)\right)_{i,j\in \{1,\hdots,n\}^2}
		\end{equation}
		
		This state representation has three advantages; it allows us to have a human-understandable input to our model, represent the surroundings of the current iterate and can approximate the inputs taken by gradient descent and Newton descend via finite difference. Another advantage is that pre-filtering can be efficiently applied by convolutions. For this state representation to represent compactly various functions, independent of their magnitude, we linearly rescale it to take its values in $[0,1]$.
			
		It is important to note that such an input becomes extremely expensive to compute as the dimensionality of the problem grows, as the size of $s_\delta^n(\cdot)$ grows exponentially with $d$. Therefore, we will use this solution for $d=2$, and discuss different ways of scaling to higher dimension in \ref{sec::highd}. 
			
		One could argue that using noisy zeroth order oracle for optimization is uncompetitive compared to higher order methods. Indeed, the study of convergence rates and lower bounds for convex optimization problem show the superiority of first-order oracles over single zeroth-order function evaluation (\cite{nemirovskii1983problem}). However, it was proven in \cite{duchi2015optimal} that by using two function evaluations, the oracle complexity of the latter type of algorithms could compete with the former, up to a low-order polynomial of the dimension factor (in the convex case). Because we use many of such samples, we are confident that we will be competitive against higher order oracles.
			
	\subsubsection{Learning the update direction/angle}
	\label{sec::angle}
	The first step of our three-step optimizer is to determine a good direction of update, given a grid of samples $s_\delta^n(f,\theta)$. In light of the previous discussion, we decided to learn this angle prediction by \emph{imitation learning}, provided two teachers: gradient descent and Newton descent. The field of imitation learning is large, though dominated by two antagonist approaches: \emph{behavioral cloning} and \emph{inverse reinforcement learning}. The latter recovers the cost function that a teacher or expert is minimizing, while the former involves training a complex model (usually a deep neural network) in a supervised fashion so that it mimics a teacher. Thorough details on both these methods, as well as a complete survey of the field of imitation learning can be found in \cite{Billard2016}. Behavioral cloning, while being straight forward and simple to implement, is known to require a large amount of data and to be prone to \emph{compounding errors}, leading to divergence between the teacher's and the imitation followed paths. 
		On the other side, inverse reinforcement learning allows the imitator to interact with the environment, and fit its behavior over whole trajectories (therefore is not affected by the compounding error issue). However, it often implies using reinforcement learning in a inner loop, making this technique rather costly to use.  In our set-up, we decided to use a behavioral cloning approach. We can indeed easily generate large amounts of training data, and are not trying to fit the entire teacher behavior but only a subpart - the direction, not the step-size.
			
	We collect our training data by launching optimization runs, where we follow the best out of the teachers (here best means leading to the largest decrease of the objective function). At each step, we record the local grid sample with a pre-determined resolution, as well as two opposite directions of update: the optimal one $d_*$ (given by the best teacher) and a set of opposite randomly generated ones $\tilde{d_*}$ (sampled to lie in the half-space defined by $\{ d\, \vert \, d^Td_* <0\}$). The expert move $d_*$ and its negative counterparts $\tilde{d}_*$ are normalized to create two actions $a_*= d_* / \lVert d_*\rVert_2$ and $\tilde{a}_*$ similarly. We then create two state-action pairs with respective label $t=1$ and $t=0$, corresponding to a positive and a negative sample. In practice, we sampled $5\cdot 10^4$ functions from $\mathcal{F}_{train}$ and let the optimization procedure run for $10$ steps on each functions, creating $10^6$ (state,action,label) tuples to train on, stored in $\mathcal{D}_{train}$.
			
	To fit the resulting (state-action) pairs, we design a simple neural network made of two convolutional layers followed by two fully connected layers. The idea of using convolutional layer is related to the problem of filtering we mentioned earlier, and to the idea that the learnt filters in the convolutional layer can act as identifiers for the different landscapes encountered during an optimization run. For a given grid of sample $s$, we denote $y(\omega,s)$ the output of this model, parametrized by the weights $\omega$ of the network. We use batch-normalization layers after the convolutional layers, and train the model to minimize the cross-entropy loss:
			\begin{equation}
				J(\mathcal{D}_{train},\omega) = \sum_{(s,a,t)\in\mathcal{D}_{train}} \Big[t\log{\{\sigma(y(\omega ,s),a)\}} + (1-t)\log{\{1-\sigma(y(\omega,s),a)\}}\Big]
			\end{equation}
			with $\sigma(y(s),a) = (1+e^{-y(\omega,s)^Ta})^{-1}$.
			The objective is to learn to correlate the output of the model when the action $a$ has a positive label (it was sampled from the best teacher). The idea of storing negative versions of that optimal action can be understood as negative sampling, or noise contrasted estimation \cite{gutmann2010noise}. We found that this approach, over the other ones we tried, lead to better performances while greatly reducing overfitting. Because we only want to use this model as an angle predictor, we will use a normalized version of the output: $\Delta(s) = y(s) / \rVert y(s) \lVert_2$.
	
		\subsubsection{Learning the step-size and the resolution}
		\label{sec::rl_reso}
		
		 At this point, we have learnt a good angle predictor. We now want to learn two new behaviors: the step-size to apply to the update, as well as the next resolution of the sample grid. Learning the step-size is obviously crucial for the optimization step. Learning the resolution is also extremely important: far from an optimum, we'd like to zoom-out to get a better understanding of the landscape. Close to an optimum, we expect an efficient system to zoom in to refine its estimation of the localization of the optimal point. Those two behaviors can't be learnt efficiently from a teacher (line-search is an unfairly good teacher for the step-size, and we simply don't have available a hand-designed teacher for the resolution).
			
		\paragraph{Reinforcement learning preliminaries} Reinforcement learning is a framework in which an agent learns its actions from interaction with its environment. The environment generates scalar values called rewards, that the agent is seeking to maximize over time. The environment is modeled as a Partially Observable Markov Decision Process (POMDP), defined to be the tuple $\left(\mathcal{O},\mathcal{S},\mathcal{A},p_0,p,q,r\right)$ where $\mathcal{O}$ is the set of observations, $\mathcal{S}$ the set of states and $\mathcal{A}$ the set of actions. $p_0(s)$ is the initial probability distribution over the states, $p(s'\vert s,a)$ the transition model, $p(o\vert s)$ the distribution of an observation conditionally to a state and $r:\mathcal{S} \to \mathbb{R}$ a function that assigns a reward to each state. The objective is to learn a policy $\pi(a\vert s) : \mathcal{S} \to \mathcal{A}$ providing the probability of choosing action $a$ in state $s$. This policy should maximize the discounted expected return $\bar{R}$:
			\begin{equation}
				\bar{R} = \E[\rho]{\sum_{t=0}^T \gamma^t r(s_t)}
			\end{equation}
			where $\gamma \in (0,1)$ is a discount factor allowing the agent to be more sensitive to rewards it will get in a close future, and the expectation being taken with respect to the state-action distribution $\rho$. A complete introduction to the reinforcement learning framework can be found in \cite{sutton1998reinforcement}.
			
		Policy search is a family of algorithm that directly search in the policy space for $\pi^* = \argmax{\pi}{\bar{R}}$. To make this search tractable, $\pi$ is usually tied to some parametrized family. A popular algorithm to perform that search is the Deterministic Policy Gradient \cite{silver2014deterministic} where we learn a deterministic parametrized policy $\pi_\eta(a\vert s) = \mu(\eta,s)$ in a fully observable Markov decision process ($\mathcal{O}=\mathcal{S}$). The system is composed of two entities, an actor and a critic. The critic, parametrized by $\omega$, has the role to evaluate the Q-values (expected return when taking an action $a$ in state $s$) of the current policy induced by the actor (parametrized by $\eta$). As it is common in actor-critic approaches, the critic is updated by batch of logged experience to minimize the squared temporal difference (TD) error $\left(r_t + \gamma Q_\omega(s_{t+1},a_{t+1}) - Q_\omega(s_t,a_t)\right)^2$. The actor's parameters are updated in the direction that maximizes the Q-values for a batch of logged states: $\Delta \eta \propto \nabla_a Q_\omega(s,a)^T \nabla_\eta \mu(\eta,s)$. \cite{lillicrap2015continuous} applied this algorithm to deep neural networks as function approximators, using techniques that were proven successful in deep Q-learning \cite{mnih2015human}, like target networks and experience replay. \cite{heess2015memory} also extended this approach for POMDP, where it is useful to use Recurrent Neural Networks as models for the policy.
			
		\paragraph{Reinforcement learning formulation} We consider the following environment for our problem. Let, for a given loss function $f$, the full state space $\mathcal{S}_f = \left\{ \theta, \alpha, \delta \right\}$ and the observations $\mathcal{O}_f = \left( s_\delta^n(f,x) \right)$. The agent hence only has access to the current grid of samples around $\theta$ with resolution $\delta$, but not to the current iterate position $\theta$, the current-step size $\alpha$ or the current resolution $\delta$. The idea behind this is to be able to generalize to unseen landscapes, and be robust to transformations such as rescaling or translations. The only events that should impact the agent's behavior is a sharp change in the neighboring landscape around the current iterate. The action space is set to be $\mathcal{A} = \{ \Delta \alpha , \Delta \delta \}\subseteq [-0.5,1]^2$ which constitutes the update rate of the step-size and the resolution. We consider deterministic transitions:
			\begin{equation}
				\begin{aligned}
					&\theta_{t+1} = \theta_{t} + \alpha_t \Delta(s_{\delta_t}^n(f,\theta_t)) \\
					&\alpha_{t+1} = \alpha_t(1+\Delta \alpha_t) \\
					&\delta_{t+1} = \delta_t(1+\Delta\delta_t)
				\end{aligned}
				\label{eq::transitions}
			\end{equation}
			where the current iterate $\theta_t$ is updated along the direction $\Delta(s_{\delta_t}^n(f,\theta_t))$ with step-size $\alpha_t$. 
			
			We have several options for the reward function. One possibility is be to consider a budgeted optimization scheme, with reward $r(s) = -f(\theta_t) \mathds{1}_{t=T}$ (the reward is only given by the final value of the function at the last step). In this case, the reward is rather sparse, and leads the trajectory search in ambiguous ways. We can prefer another solution, where the whole trajectory of the agent over the landscapes is evaluated: $ r_f(s_t) = -f(\theta_t)$. 
			This leads the policy search to optimize for the following return:
			\begin{equation}
				R_f = -\sum_{t=0}^T \gamma^t f(\theta_t)
				\label{eq::reward}
			\end{equation}	
			Note that for $\gamma = 1$ this leads us to optimize over the same criterion that \cite{li2017learning} and \cite{andrychowicz2016learning} (in the former, the authors call this the meta-loss). 
			
		It is important to note that the previously described POMDP, that we will denote as $\mathcal{M}_f$, is parametrized by a function $f$ sampled inside $\mathcal{F}_{train}$. This induces a distribution $\mathcal{M}_{train}$ over POMDPs. In the following experiments, we won't make that distinction and train a single parametrized policy on the resulting POMDP distribution - that implies that every new episode is generated with $\mathcal{M}_f \sim \mathcal{M}_{train}$. This induces a difficulty over the learning task: both the transitions and the reward defined in \eqref{eq::transitions} and \eqref{eq::reward} change between every episode. To help the agent figure out optimal moves, we can change the reward so that it becomes insensitive to the magnitude of the sampled function $f$ and the position of the initial iterate $\theta_0$:
			\begin{equation}
				r_f(s_t) = -\frac{f(\theta_t) - f(\theta_f^*)}{f(\theta_0)-f(\theta_f^*)}
			\end{equation}
			with $\theta_f^* = \argmin{\theta}{f(\theta)}$. To also help the agent optimize over long trajectories where the magnitude of $f(\theta_0)$ largely surpasses $f(\theta_f^*)$, we propose a second version of the reward function:
			\begin{equation}
				r_f(s_t) = -\frac{f(\theta_t) - f(\theta_f^*)}{\bar{f}_k-f(\theta_f^*)}
			\end{equation}
			with $\bar{f}_k$ being the mean value of the objective function over the last $k$ iterates (we found that in our set-up, $k=5$ provides good results). The use of this reward function was a crucial element in the success of our reinforcement learning approach.
			
	We model the agent policy by a recurrent neural network, made up of two convolutional layers, followed by a Long-Short Term-Memory cell (LSTM, introduced by \cite{hochreiter1997long}), followed itself by two hidden layers. The critic is modeled by a similar network, and both were trained using the DPG algorithm. During training, we sample $f\sim\mathcal{F}_{train}$ at the beginning of each episode. The initial iterate is randomly sampled in the landscape so that it is far away enough from the optimum of the loss function. The episode is ran for a fixed horizon $T=30$ and we fix the discount factor $\gamma$ to 1. 
	
\subsection{Architecture for $d>2$}
\label{sec::highd}

	The idea of using grid samples $s_\delta^n(f,\theta)$ can't be exploited in high-dimensional problems as its size grows exponentially with $d$. To extend our framework for $d>2$, we consider the following set-up: let $f: \mathbb{R}^d \to \mathbb{R}$ and $\theta$ the initial iterate. We note $s(f,\theta,i,j)$ the vector that contains the two-dimensional grid sampled at $\theta$ along the dimensions $i$ and $j$. In other words, with $\delta_i$ the $d$-dimensional vector whose entries are all 0 but the ith one that is set to 1, and $E_{i,j} = (\delta_i,\delta_j)$ a $d\times2$ matrix, we note: $s_\delta^n(f,\theta,i,j) = s_\delta^n(f,E_{i,j}^T\theta)$.
		
		By considering all pairs of dimensions, we can compute $d(d-1)/2$ of such grids, leading to the prediction of as many angles $\Delta_{i,j}(f,\theta) = \Delta(s_\delta^n(f,\theta,i,j))$, step-size updates $\Delta \alpha_{i,j}$ and resolution updates $\Delta \delta_{i,j}$ - all predicted with the models trained in the two dimensional case. Therefore, if we keep record of all step-size and resolution for every pair of dimension $(i,j)$ we can compute $d(d-1)/2$ updates $\Delta \theta_{i,j} = \alpha_{i,j} \Delta_{i,j}(f,\theta)$. We can consider each of these outputs like the $d(d-1)/2$ two-dimensional projections of the true $d$-dimensional update $\Delta \theta$ so that $\Delta \theta_{i,j}= E_{i,j}^T \Delta \theta$. We can therefore try to retrieve $\Delta\theta$ by a least-square approach, and find $\Delta \hat{\theta}$:
		
		\begin{equation}
			\Delta \hat{\theta}\triangleq \argmin{\delta \theta}{ \sum_{1\leq i < j \leq d} \left(E_{i,j}^T \delta\theta - \Delta\theta_{i,j}\right)^2}
		\end{equation}
		
		Solving this equation leads to the analytical expression:
		\begin{equation}
			\begin{aligned}
				\Delta \hat{\theta}
							&= \frac{1}{d-1} \sum_{1\leq i < j \leq d} \alpha_{i,j} E_{i,j}\Delta_{i,j}(f,\theta)
			\end{aligned}
		\end{equation}
		
		Each pair of dimension has a corresponding step-size $\alpha_{i,j}$  and resolution $\delta_{i,j}$, which are updated by running the associated two-dimensional grid through the system described in \ref{sec::rl_reso}. This computation requires maintaining $d(d-1)/2$ learning rates and resolutions, and computing as many grid samples. Because of this quadratical growth with the dimension, this leads to clock-time and memory issues for large values of $d$. A simple way round this problem is to sample $k<d$ pairs of dimensions, compute $\Delta\hat{\theta}$ based only on these $k$ pairs and update their corresponding learning rates and resolutions. If we note $\Theta_k$ the set of $k$ pairs we sampled:
		
		\begin{equation}
			\Delta \hat{\theta}_k =  \frac{1}{k-1} \sum_{(i,j)\in\Theta_k} \alpha_{i,j} E_{i,j}\Delta_{i,j}(f,\theta)
			\label{eq::d_update}
		\end{equation}
		Different strategies can be employed to sample $\Theta_k$, possibly leveraging some knowledge about the optimization problem's structure. Such strategies are experimentally investigated in \ref{sec::net_exp}.
				

\section{Results}
{
	\label{sec::experiments}
	\subsection{Behavioral analysis}
	{
		
		\paragraph{Angle predictor} The training of the angle predictor is straight forward and leads to robust angle prediction. Table \ref{tab::angle_result} shows the mean angle dissimilarity between the learnt angle predictor and the best teacher (as defined in \ref{sec::angle}) on a set of held-out functions from $\mathcal{F}_{train}$. We show these results when training and testing on either single modalities (e.g only quadratic, only valleys, ..) of $\mathcal{F}_{train}$, or all of them at the same time. We see that training the angle predictor on the whole meta-dataset does not impact its predictions abilities compared to landscape-specific training. Indeed, we only see a small drop in the quality of the predictions, which we attribute to the partial observability of the local landscapes (a valley seen under a small resolution can locally appear like a quadratic bowl, for instance). 
			\begin{table}[h!]
				\centering
				\begin{tabular}{|c|c|c|c|c|}
				\hline
			 	\backslashbox{Train}{Test} & Quadratics & Valleys & Saddles & Plateaus+cliffs\\
				\hline
				Quadratics & 1.2 &  89.2 & 43.1 & 34.8\\
				Valleys & 83.5 & 4.9 & 94.1 & 86.9 \\
				Saddles & 44.7 & 85.8 & 3.8 & 57.2 \\
				Plateaus+cliffs & 29.4 & 94.3 & 53.9 & 1.9 \\
				\hline 
				\hline 
				All & 3.1 & 6.4 & 5.1 & 3.8 \\
				\hline
				\end{tabular}
				\caption{Mean angle dissimilarity in degrees on held-out functions from $\mathcal{F}_{train}$ when training on both single modalities and the whole meta-dataset.}
				\label{tab::angle_result}
			\end{table}
		
		\paragraph{Resolution and step-size predictor} Evaluating exhaustively the learnt policy for updating the resolution and the step-size is complicated. We provide in Appendix \ref{sec::policy_viz} the trajectories of the step-size and resolution under the learnt policy, for functions of $\mathcal{F}_{train}$. Independently of the quality of this policy (which will be evaluated in the following sections), we can notice it follows our intuition of what is a good policy in this context: it zooms out in the beginning of the iteration procedure, while zooming in near the end when the local landscapes indicates the presence of a minimum nearby. The step-size evolution follows a similar logic. We also present in Appendix \ref{sec::policy_viz} some visualization of the dynamics when the initial state (i.e initial learning or resolution) is purposely set to a misleading value, and show how the policy recovers from this poor initialization. 
	}
	\subsection{Two-dimensional experiments}
	{
		We first want to evaluate the optimizer resulting from the model we introduced on $\mathcal{F}_{train}$ to evaluate its behavior on known landscapes. This already can be seen as some kind of meta-testing on some hold-out since we only sample in $\mathcal{F}_{train}$, which contain an infinite number of functions. Therefore, we can assume that whenever we sample in $\mathcal{F}_{train}$, we will obtain a function that the optimizer has not seen during training. 
		
		
		\begin{figure}[h!]
			\centering
			\begin{subfigure}[b]{0.45\linewidth}
			{
				\centering
				\includegraphics[width=0.8\textwidth]{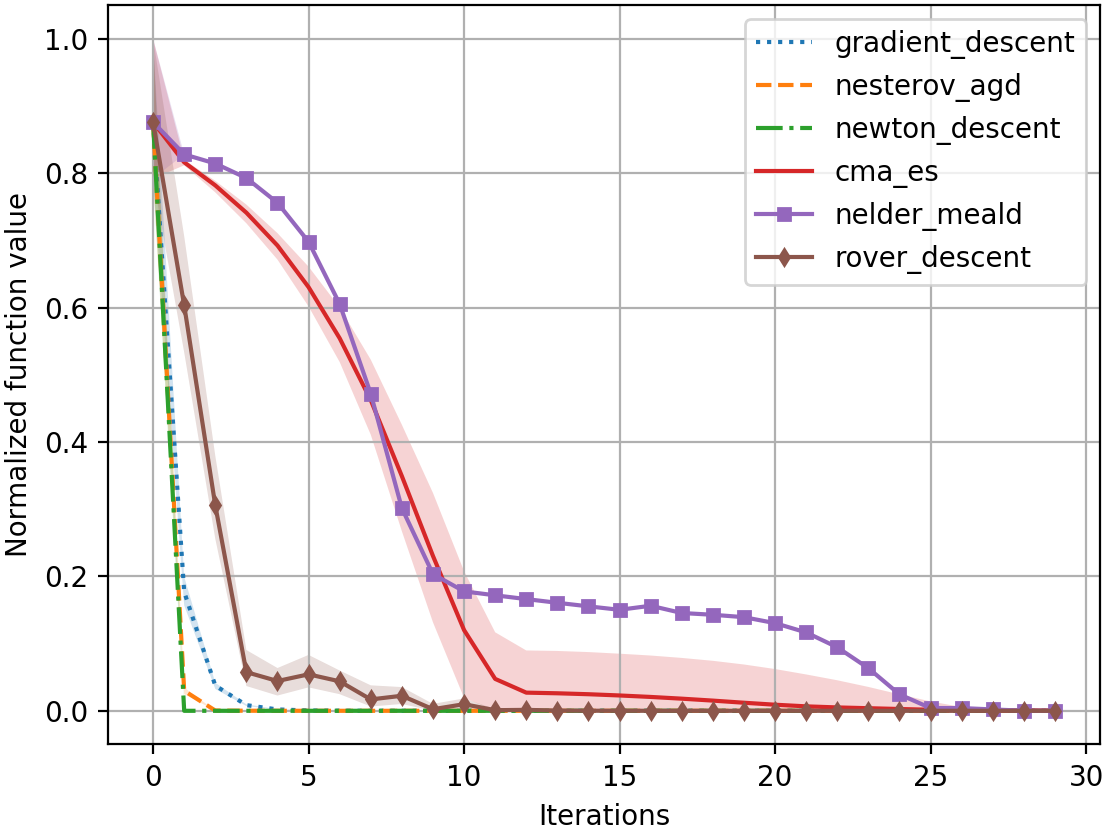}
				\caption{Quadratic}
				\label{fig::fquad_eval}
			}
			\end{subfigure}
			\begin{subfigure}[b]{0.45\linewidth}
			{
				\centering
				\includegraphics[width=0.8\textwidth]{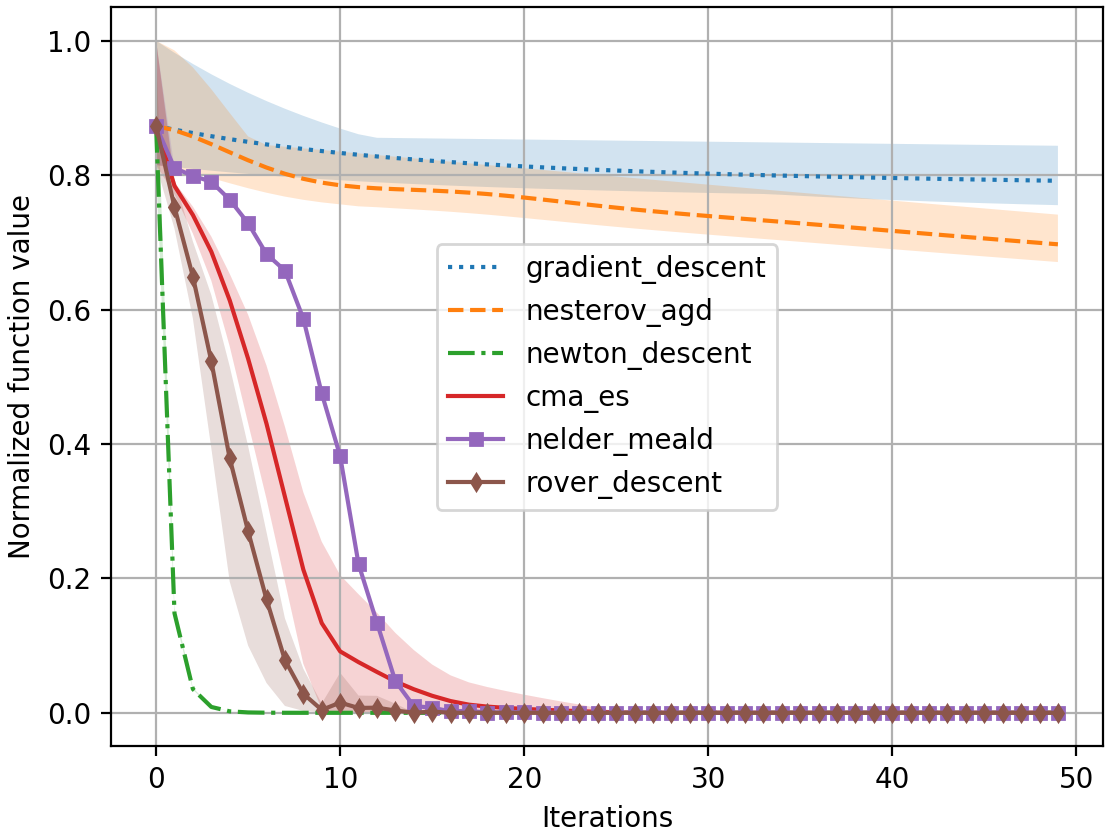}
				\caption{Valley}
				\label{fig::valley_eval}
			}
			\end{subfigure}\\
			\begin{subfigure}[b]{0.45\linewidth}
			{
				\centering
				\includegraphics[width=0.8\textwidth]{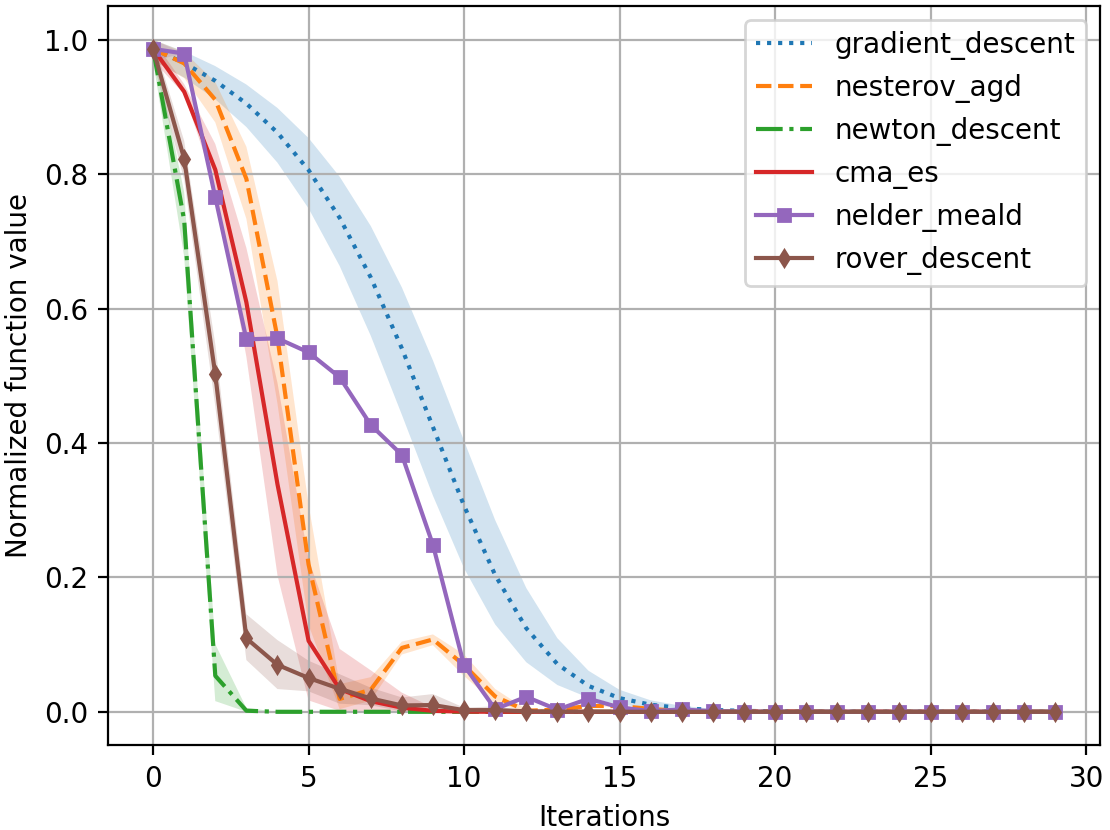}
				\caption{Plateau+cliff}
				\label{fig::uni_eval}
			}
			\end{subfigure}
			\begin{subfigure}[b]{0.45\linewidth}
			{
				\centering
				\includegraphics[width=0.8\textwidth]{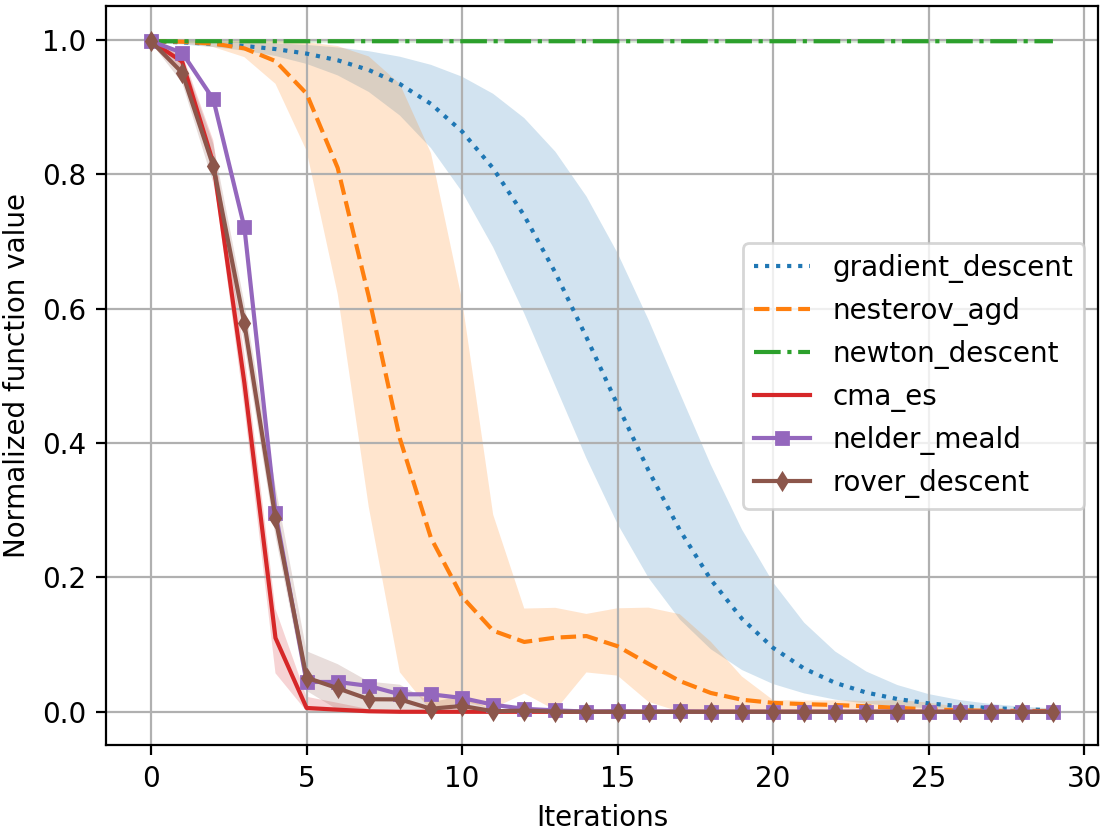}
				\caption{Saddle}
				\label{fig::saddle}
			}
			\end{subfigure}
			\caption{Tests runs on instances of modalities of $\mathcal{F}_{train}$. The shades represents the envelope of the trajectories over an entire 20-fold. Best viewed in color.}
			\label{fig::ftrain_test}
		\end{figure}
		
		We follow a simple procedure: we sample $f\in\mathcal{F}_{train}$, an initial point $\theta_0 \in \Omega_0$, and sample uniformly at random an initial step-size and an initial resolution inside the distribution used at meta-training time. We then add a small perturbation to the initial iterate and run an optimization trajectory with a fixed horizon. We repeat this procedure many times to evaluate the global sensitivity of our algorithm to the position of the first iterate. To compare its performance with a broad variety of optimizers, we decide to evaluate with the same procedure a collection of optimization algorithms that include: gradient descent, Nesterov accelerated gradient descent, Newton Descent, Covariance Matrix Adaptation Evolution Strategy (CMA-ES) and the Nelder-Meald method. The results are regrouped in Figure \ref{fig::ftrain_test}. The lines represent the mean trajectory of each optimizer, while the shaded areas represent the envelope of all its trajectories (that were generated from noisy versions of the initial iterate).  The results, shown here for a single function $f$ and iterate point $\theta_0$ are consistent in our experiments: we have learnt to compete with a wide variety of hand-designed algorithms. The hyper-parameters of hand-designed optimizers are modified at each time to perform as well as possible on the whole modality of $\mathcal{F}_{train}$ we are testing on. This means that our learnt optimizer sometimes compete with unfairly good algorithms (like Newton descent on a quadratic loss, that hits the optimum after just one iteration). In some cases, the apparent lack of trajectory envelope is due to the fact that the perturbation on the initial point sometimes have to be reduced for visualization purposes. 
		
		To evaluate the meta-generalization abilities of our learnt optimizer, we also evaluate it on a two-dimensional meta-testing dataset $\mathcal{F}_{test}$. We selected various two-dimensional optimization problems known to be challenging for general optimization methods. The complete list contains Rosenbrock, Ackley, Rastrigin, Maccornick, Beale and Styblinksi's function, for which the literal expressions and surface plots can be found in Appendix \ref{sec::ftest}. It is important to note that none of these landscapes were seen by the optimizer during its training. 
		
		For each of those functions, we select a starting point that constitute a challenge for all compared optimizers (also indicated on the surface plot in Appendix \ref{sec::ftest}). We then followed the previously described procedure, and set the hand-designed optimizers hyper-parameters to show good behavior for every small perturbation of the initial iterate. The results are displayed in Figure \ref{fig::ftest_test}, and remains consistent when changing the starting point for each of the meta-test function. Our learnt optimizer can generalize to new landscapes, even multimodal ones, and compete with a wide variety of optimizers. On these multimodal landscapes, CMA-ES and the Nelder-Meald method provide two strong baselines (only the mean trajectory appear for such functions for vizualisation purposes). On \ref{fig::ackley_eval}, they are the only two algorithms with our method that find the global optimum. However, in \ref{fig::rastrigin_eval}, only our method finds for every perturbation of the initial iterate the global minimum. 0ur optimizer starts by zooming out to get a better understanding of the landscape, leading it to quickly cover an area where lays the global minimum. 


		\begin{figure}[h!]
			\centering
			\begin{subfigure}[b]{0.3\linewidth}
			{
				\centering
				\includegraphics[width=\textwidth]{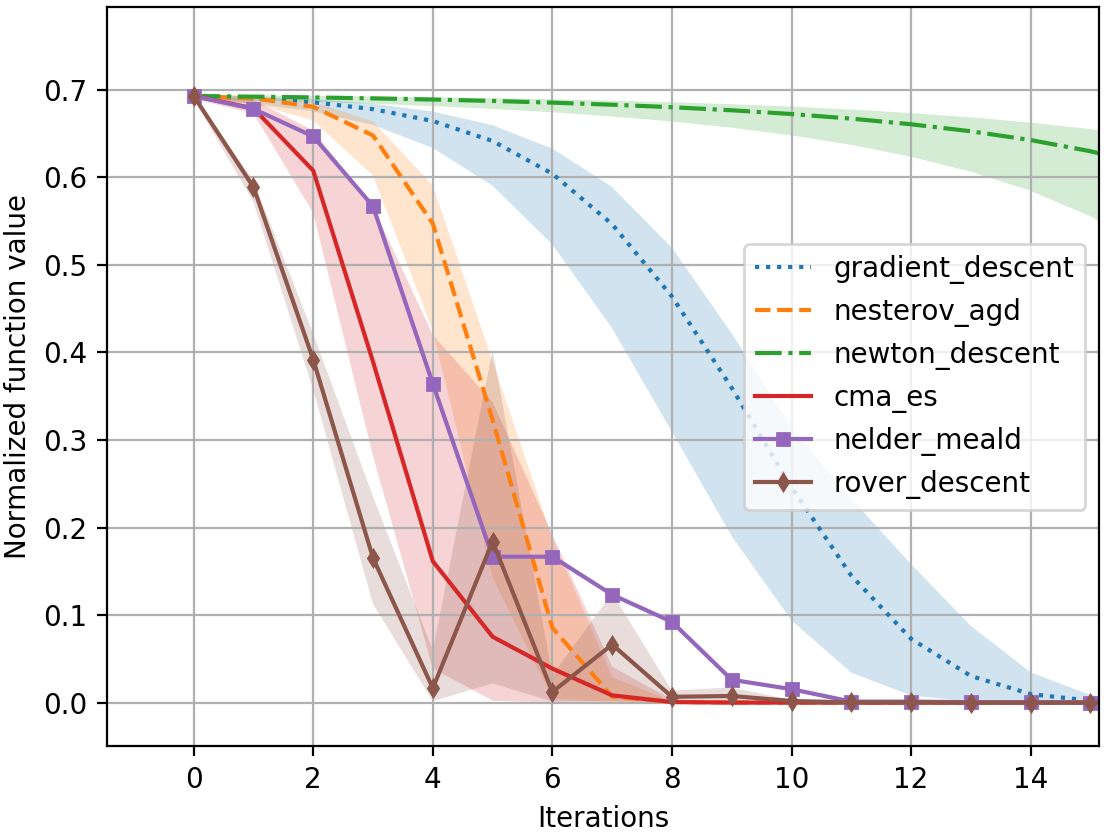}
				\caption{Styblinksi}
				\label{fig::styblinksi_eval}
			}
			\end{subfigure}\hfill
			\begin{subfigure}[b]{0.3\linewidth}
			{
				\centering
				\includegraphics[width=\textwidth]{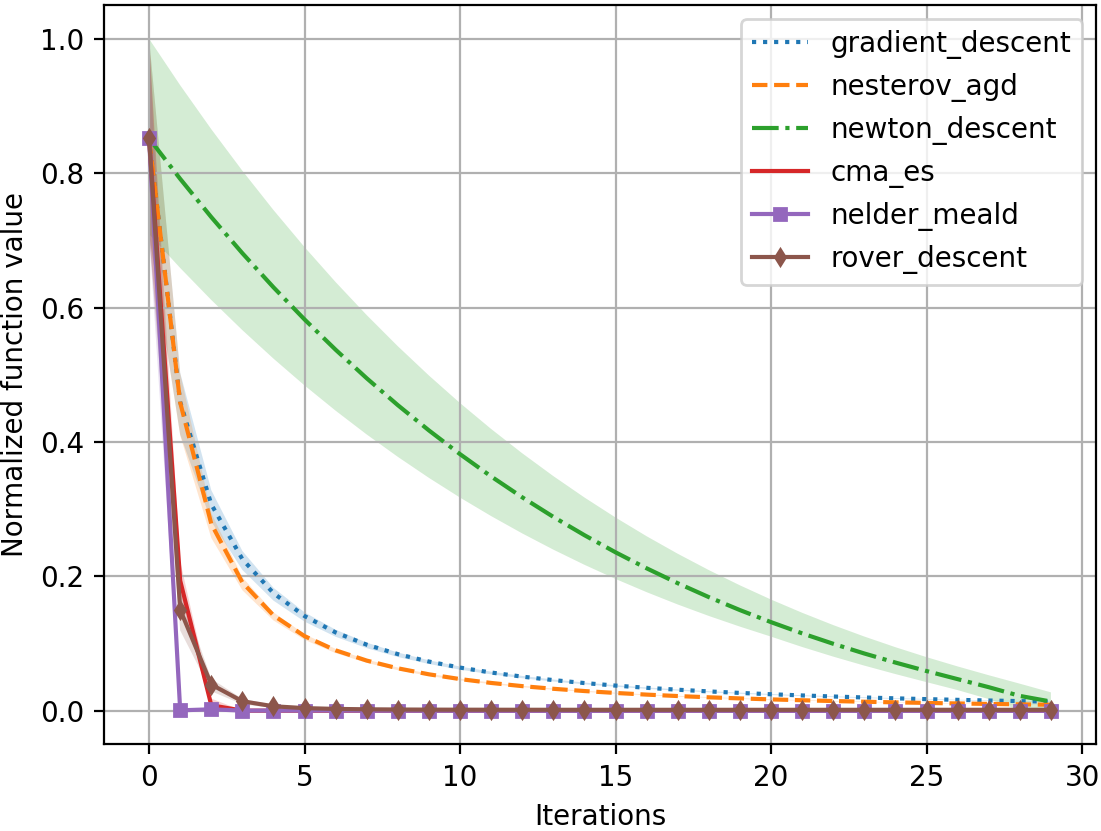}
				\caption{Beale}
				\label{fig::beale_eval}
			}
			\end{subfigure}\hfill
			\begin{subfigure}[b]{0.3\linewidth}
			{
				\centering
				\includegraphics[width=\textwidth]{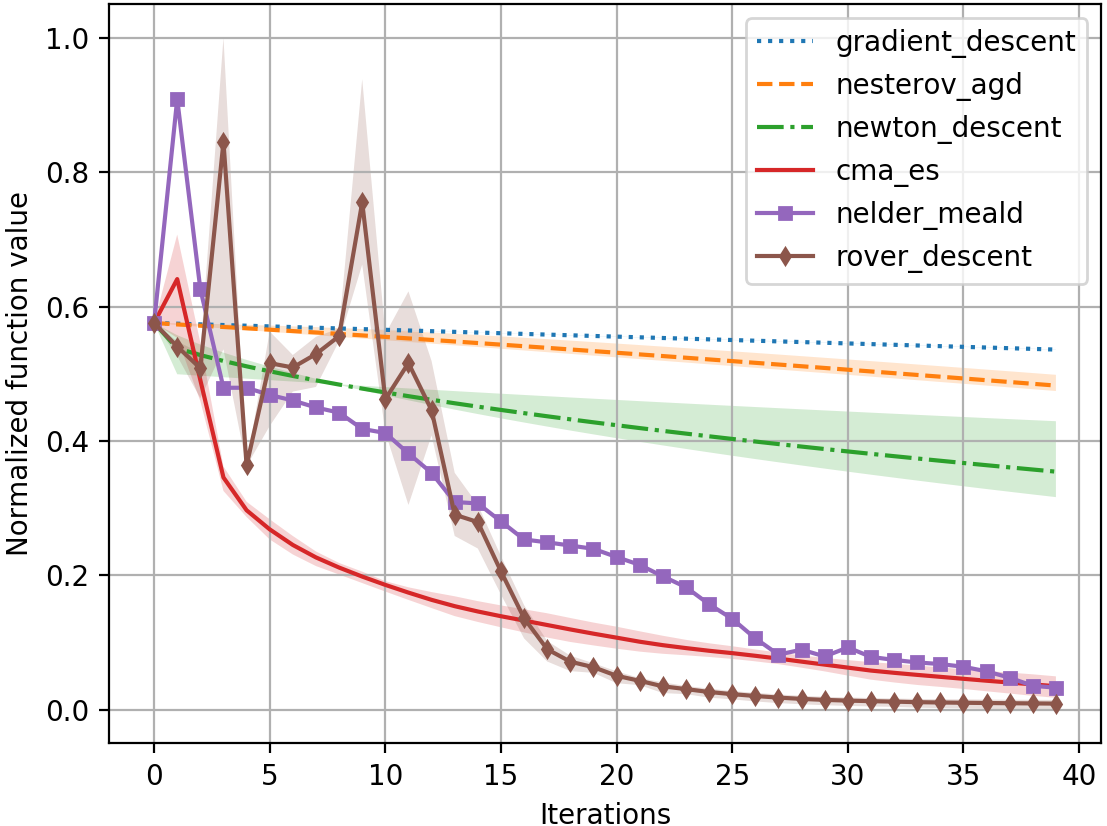}
				\caption{Rosenbrock}
				\label{fig::rosenbrock_eval}
			}
			\end{subfigure}\\
			\begin{subfigure}[b]{0.3\linewidth}
			{
				\centering
				\includegraphics[width=\textwidth]{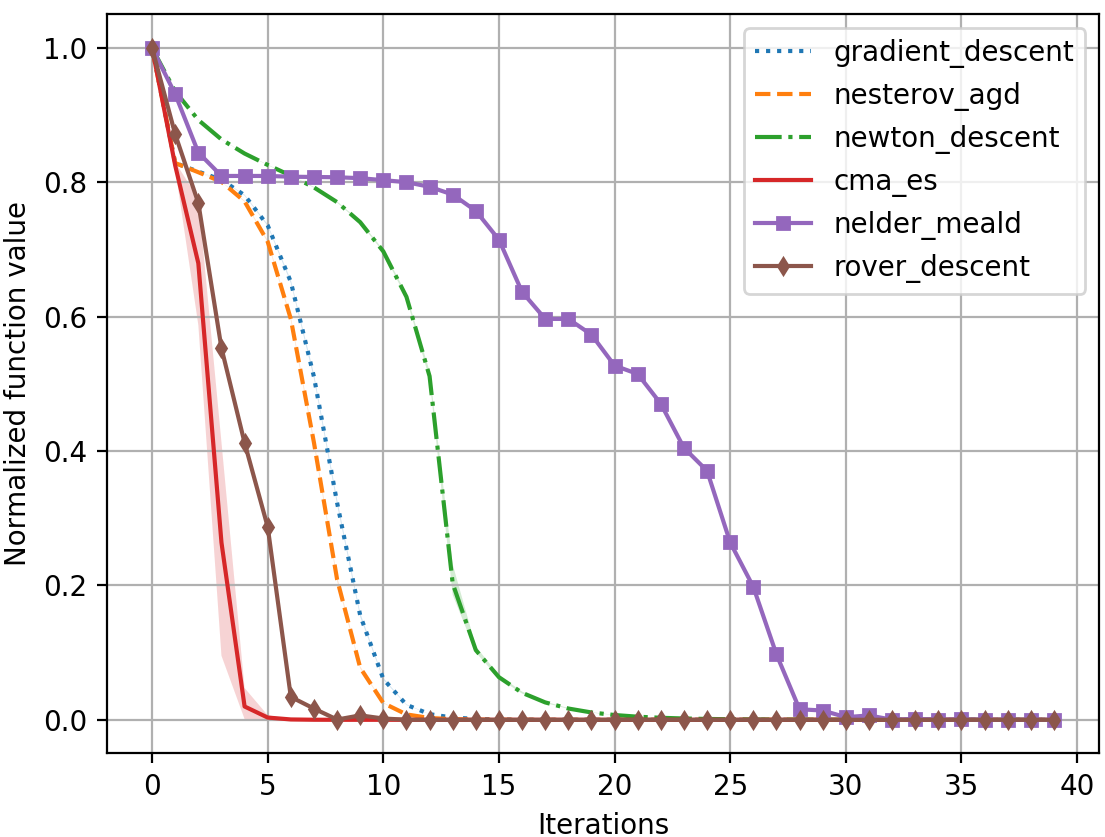}
				\caption{Maccornick}
				\label{fig::maccornick_eval}
			}
			\end{subfigure}\hfill
			\begin{subfigure}[b]{0.3\linewidth}
			{
				\centering
				\includegraphics[width=\textwidth]{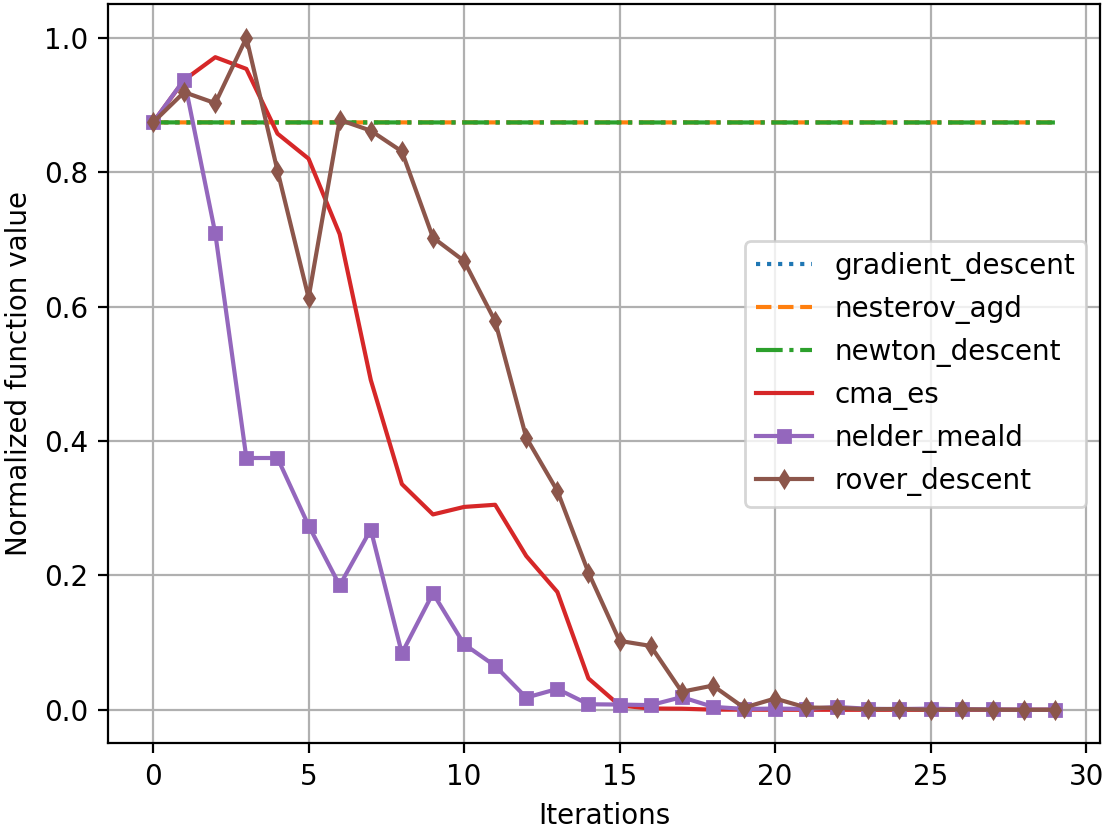}
				\caption{Ackley}
				\label{fig::ackley_eval}
			}
			\end{subfigure}\hfill
			\begin{subfigure}[b]{0.3\linewidth}
			{
				\centering
				\includegraphics[width=\textwidth]{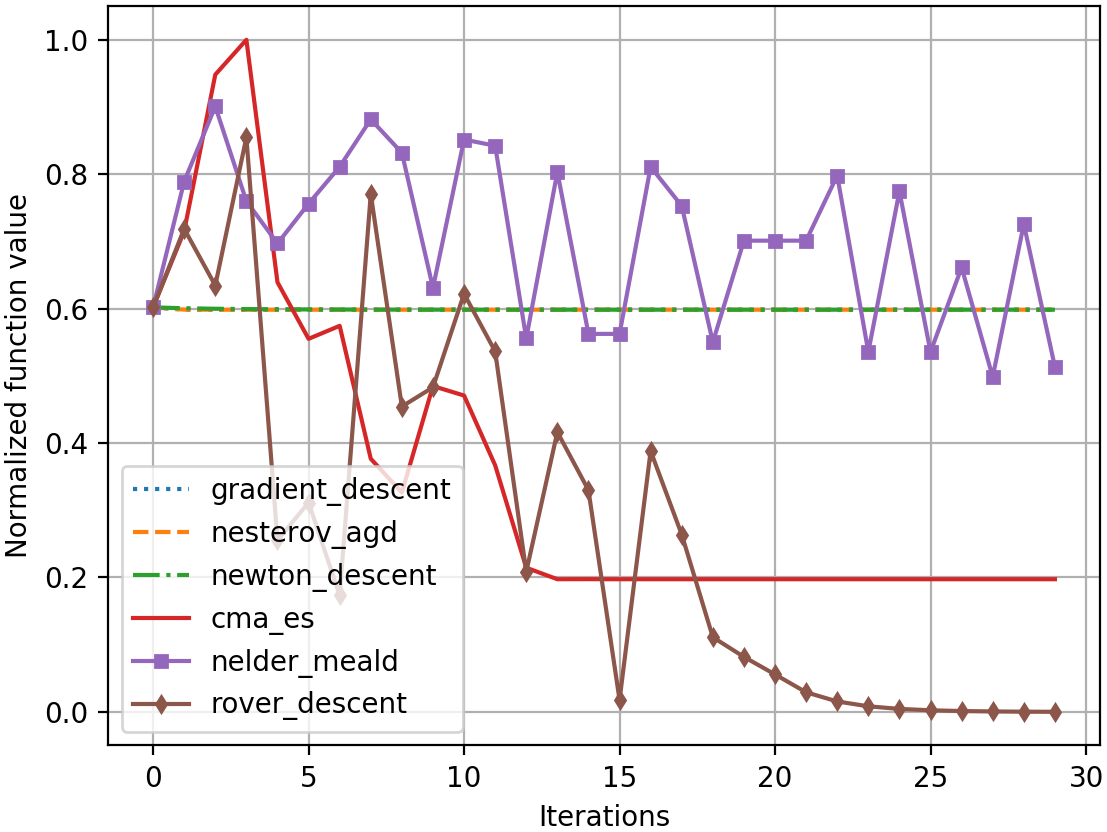}
				\caption{Rastrigin}
				\label{fig::rastrigin_eval}
			}
			\end{subfigure}
			\caption{Tests runs on instances of modalities of $\mathcal{F}_{test}$. The shades represents the envelope of the trajectories over an entire 20-fold. Best viewed in color.}
			\label{fig::ftest_test}
		\end{figure}
	}
	
	
	\subsection{High-dimensional experiments}
	{
		We now test the procedure described in Section \ref{sec::highd} for problem of dimensions $d>2$. We propose to do this by considering a linear classifier for a binary classification task, and a small neural network classifier. 
		
		\subsubsection{Linear binary classification}
		{
			We generate random binary classification tasks in dimension $d>2$, according to the framework described in \cite{guyon2003data}. We want to optimize over the cross-entropy loss induced by this dataset. We therefore sample an initial iterate $\theta_0$, an initial learning rate and an initial resolution for our optimizer, and launch an optimization run. We test against two fairly good optimizers for this task: tuned gradient descent and Newton descent. We use a fixed budget $k=10$ of dimensions we can sample at each iteration - as described in \ref{sec::highd}. The results for three different randomly generated datasets of different dimensions are presented in Figure \ref{fig::highd}. 
		
			The results presented here are consistent in our experiments: our learnt procedure competes with tuned optimizers that use respectively first and second order information. However, one major downside of our optimizer is clock-time performances - one optimization run in this simple set-up can take up to a minute for $d>50$, against a few seconds for gradient descent on the machine used for our experiments. Also, its performance is impacted by the under sampling that happens when the budget $k$ is significantly smaller than the dimension $d$. Increasing the budget $k$ improves the per-iteration performance but greatly impact the algorithm's clock-time (as the number of operations grows quadratically with $k$). In the following experiments, we propose sampling strategies for the pairs of dimensions used at every update that take advantage of the problem's structure to cope with this limitation.
		
			
			\begin{figure}[h!]
			\centering
			\begin{subfigure}[b]{0.33\linewidth}
			{
				\centering
				\includegraphics[width=\textwidth]{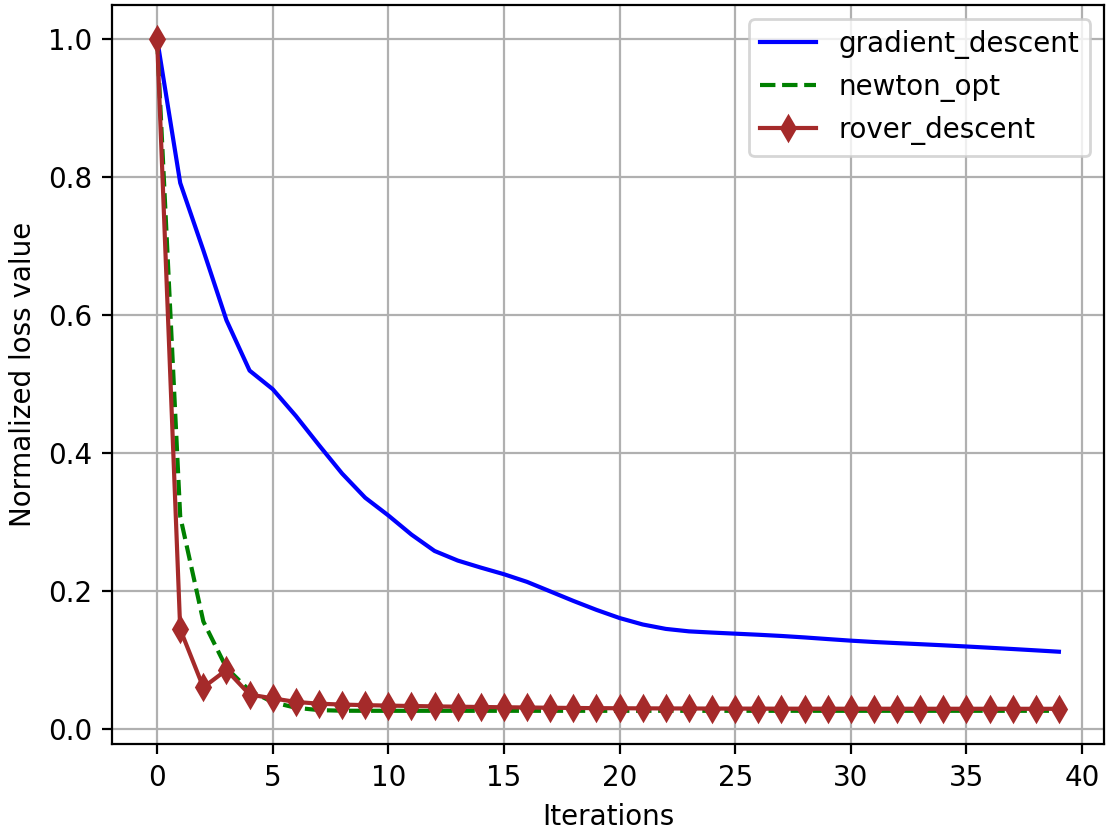}
				\caption{d=10}
				\label{fig::bc_10}
			}
			\end{subfigure}\hfill
			\begin{subfigure}[b]{0.33\linewidth}
			{
				\centering
				\includegraphics[width=\textwidth]{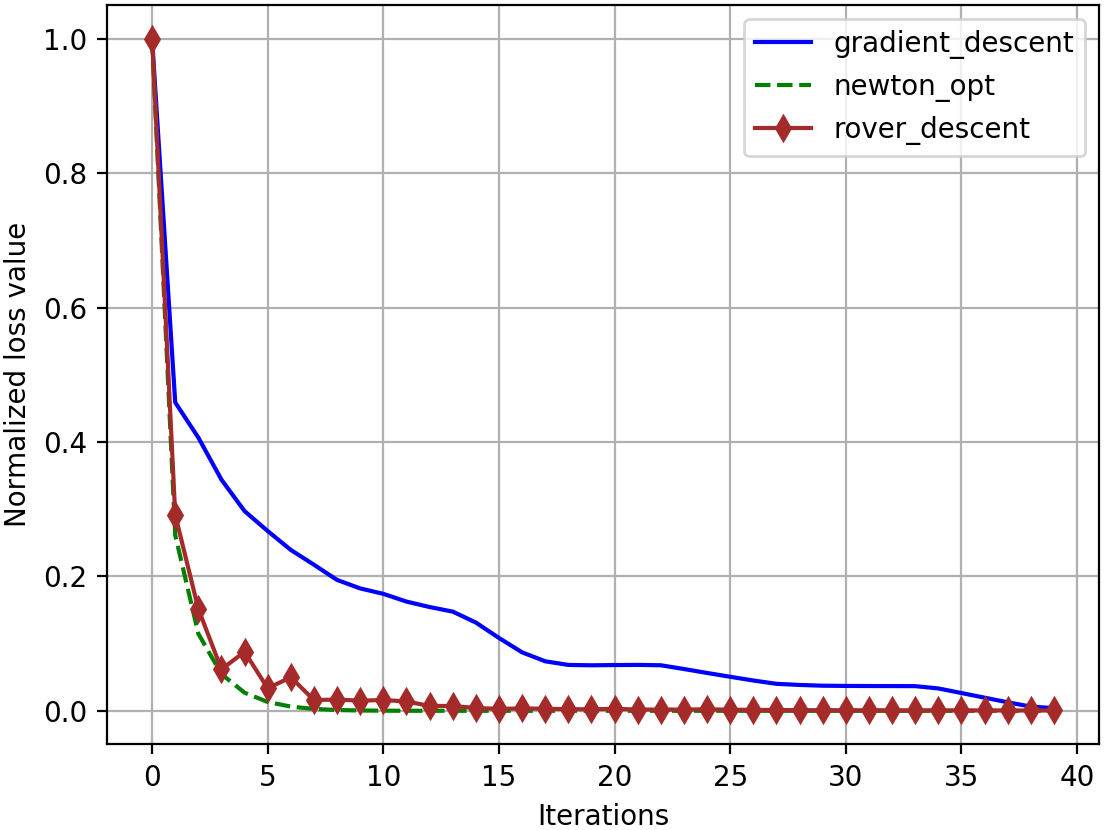}
				\caption{d=20}
				\label{fig::bc_20}
			}
			\end{subfigure}\hfill
			\begin{subfigure}[b]{0.33\linewidth}
			{
				\centering
				\includegraphics[width=\textwidth]{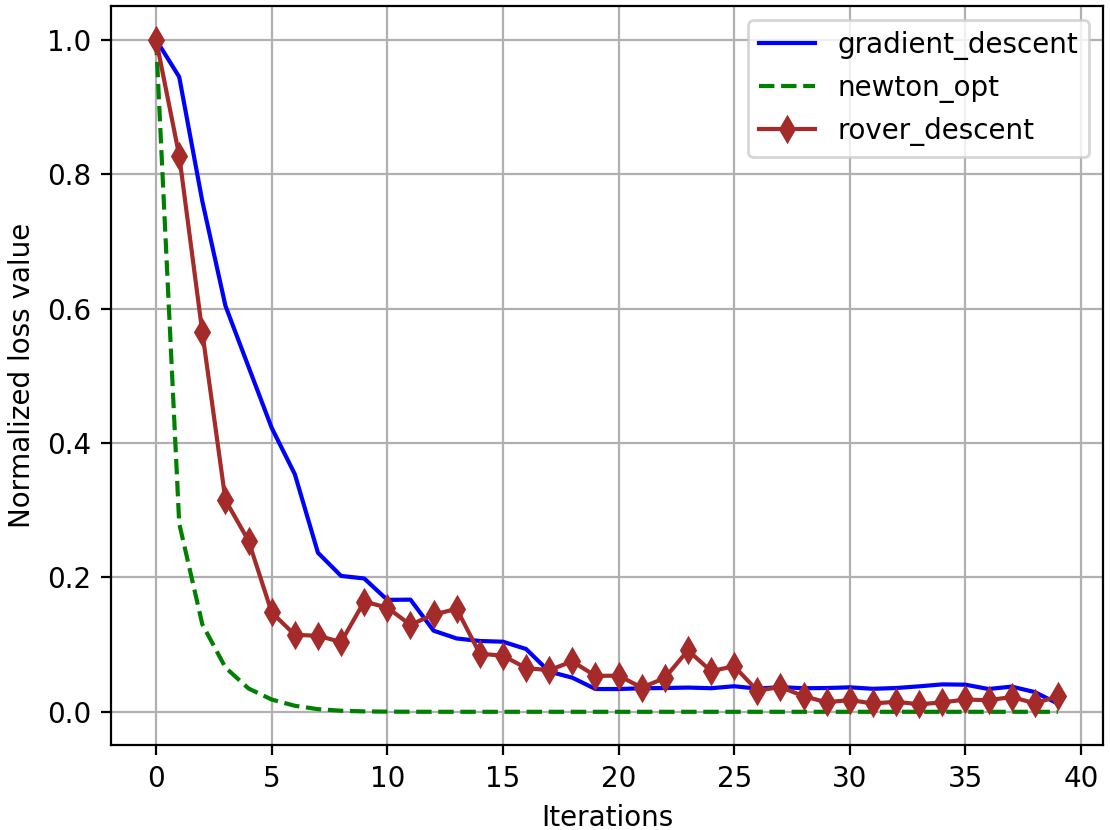}
				\caption{d=50}
				\label{fig::bc_50}
			}
			\end{subfigure}
			\caption{Test runs for cross entropy loss of randomly generated binary classifications tasks of dimension $d$, with budget $k=10$ of pair samples per iterations. Best viewed in color.}
			\label{fig::highd}
		\end{figure}

		}
		
		\subsubsection{Small neural network}
		{
			\label{sec::net_exp}
			We now want to use Rover Descent for a more complicated task. We consider using a small neural network in order to solve the Iris dataset \cite{fisher1936use}, which consists of 150 instances of 4-dimensional inputs and their respectives labels (1,2, or 3). The neural network we use is a small neural network, with two hidden-layer of width 10 and a softmax output activation, alongside a cross-entropy loss. The dimension of its loss landscape is $d=193$. We compare our results with tuned gradient descent and Adam. 
			
			Figure \ref{fig::netk} show the results we obtained using the same sampling strategy as presented earlier (\emph{i.e} we sample uniformly at random $k$ dimensions for which we create all possible $k(k-1)/2$ pairs to create two-dimensional slices). It now appears that simply increasing $k$ is not enough to ensure good behavior. Because some dimensions are visited by the algorithm only late in the procedure, their corresponding learning rate is still equal to the initial one, which can lead to erratic behaviors close to local minimum. Also, increasing $k$ implies that for every dimension we sample, we increase the number of two-dimensional moves our algorithm receives. Unlike for the convex case of linear binary classification, it appears that taking their mean value is a poor strategy as it leads to a slight decrease in performance. 
			
			To improve those results, we decide to use a slightly different sampling strategy. For \emph{every} dimension, we are going to create pairs with $l$ other dimensions, sampled uniformly at random. The difference is that now every dimension will be used at least once at every update. The number of pairs we create is now $l\times d$. Figure \ref{fig::netl} present the result obtained for different values of $l$ and proves the superiority of this approach over the previous one. 
			
			Finally, we decide to use the special structure of the neural network to improve our algorithm. We use the same sample strategy we just presented, except that now the $l$ dimensions needed for every dimensions are sampled within a pre-defined subset. More precisely, we want to leverage a block-diagonal structure of the Hessian of the neural network: for every dimension (which correspond to a weight or a bias of the neural network), we only create pairs with dimensions corresponding to a weight or bias belonging to the same layer. Figure \ref{fig::netlblock} present the results obtained, which again improves against the last one and compete with Adam on this task. Leveraging a block-diagonal approximation of the Hessian is not a new idea and was recently used to obtained state-of-the-art result on neural network optimization (\cite{martens2015optimizing},\cite{zhang2017block})
			
			\begin{figure}[h!]
			\centering
			\begin{subfigure}[b]{0.33\linewidth}
			{
				\centering
				\includegraphics[width=\textwidth]{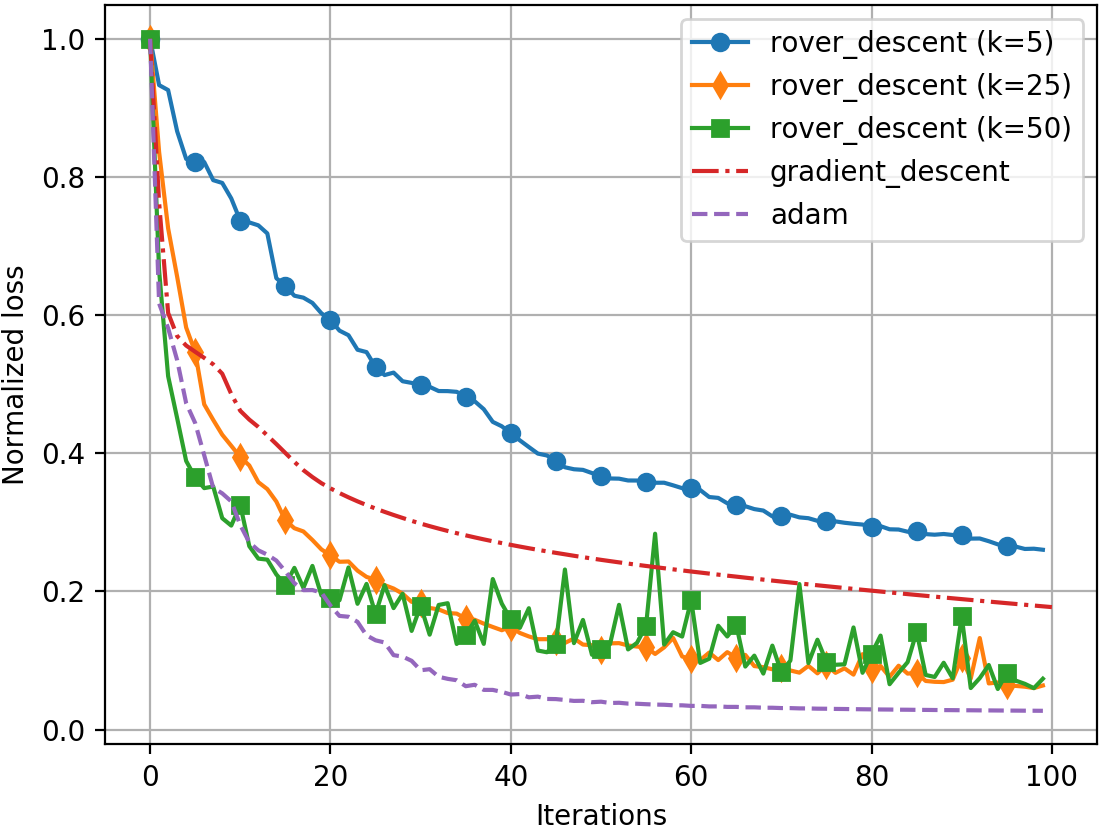}
				\caption{Pair sampling}
				\label{fig::netk}
			}
			\end{subfigure}\hfill
			\begin{subfigure}[b]{0.33\linewidth}
			{
				\centering
				\includegraphics[width=\textwidth]{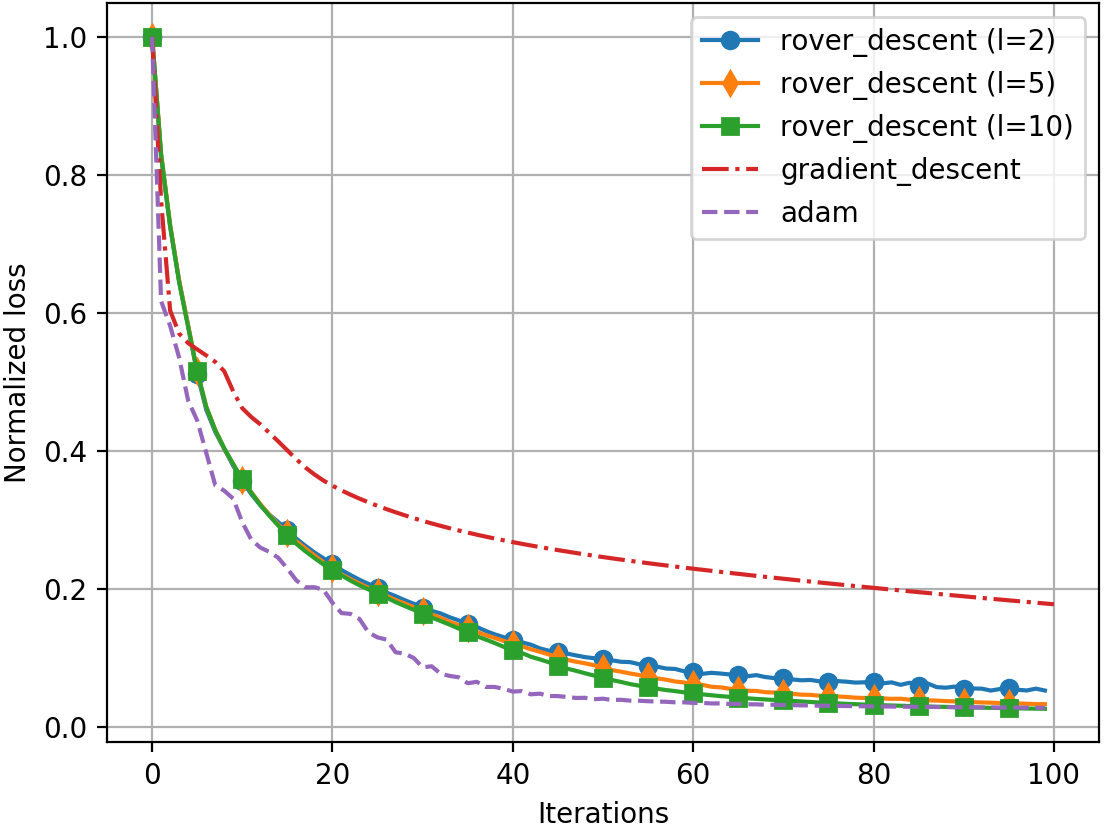}
				\caption{Per-dimension sampling}
				\label{fig::netl}
			}
			\end{subfigure}\hfill
			\begin{subfigure}[b]{0.33\linewidth}
			{
				\centering
				\includegraphics[width=\textwidth]{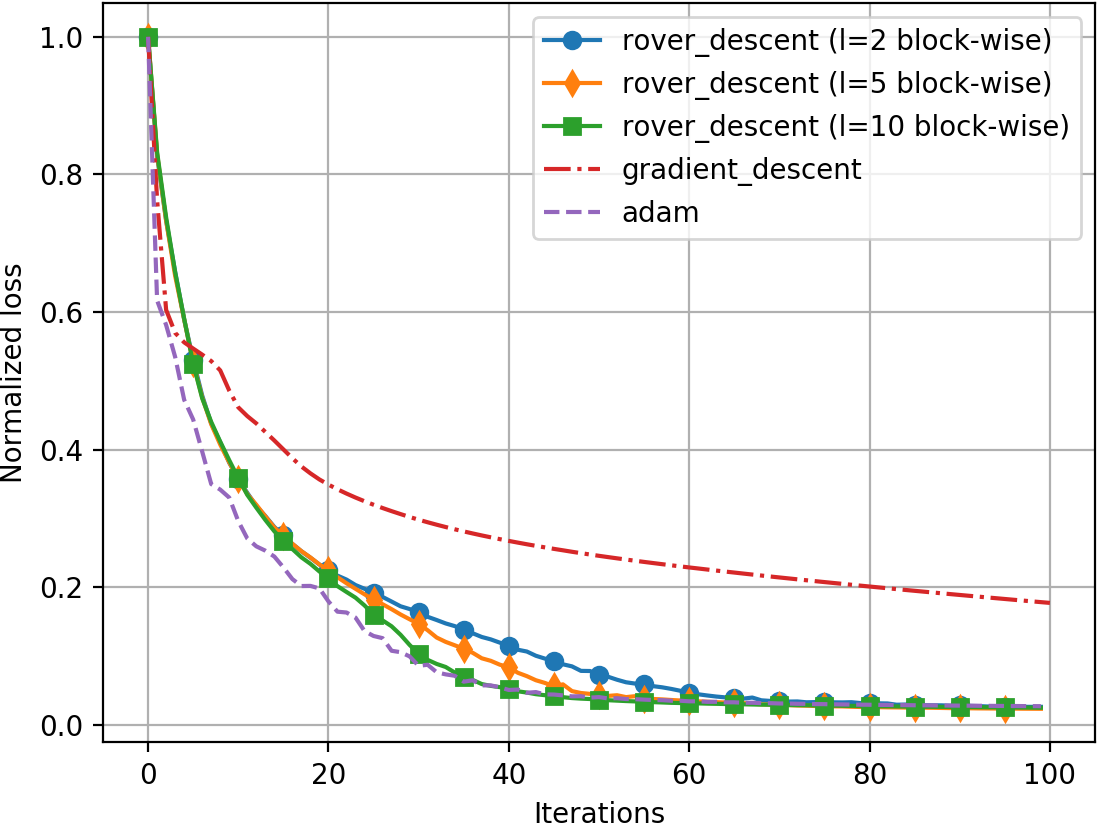}
				\caption{Per-dimension block sampling}
				\label{fig::netlblock}
			}
			\end{subfigure}
			\caption{Test runs on a small neural network for Iris dataset, with different sampling strategies for the two-dimensional slices. Best viewed in color.}
			\label{fig::net_test}
		\end{figure}

		}
	}
}


\section{Conclusion}
{

\label{sec::conclusion}
We introduced a new framework in order to achieve meta-generalization when learning to optimize. By combining tools from imitation learning and reinforcement learning, and defining the meta-dataset as a small set of prototypical functions that frequently appear in optimization problems, we were able to learn an optimization algorithm that generalizes well to unseen loss landscapes. Though this learnt optimizer is dependent on the setting of some hyper-parameters, we showed that their tuning doesn't have a meaningful impact on performance, as the learnt optimizer is quite robust and can recover from bad initializations.
   
   In future work, we plan on running our framework on more complex, high-dimensional non-convex problems - such as the training of deep neural networks, which will require some adaptations for reducing the clock-time of our optimizer. Also, we can imagine having more control on the meta-train dataset in order to scan for such complicated landscapes. For instance, following a leave-one-out procedure, we could try to estimate the prototypical landscapes that are needed by a learnt optimizer to train a deep neural network, and therefore make further assumptions over the composition of the related loss landscapes. Furthermore, we wish to develop adaptative methods of dimension sampling in order to make our optimizer more efficient in very high dimensional problems. A more thorough evaluation of its performances will then involve testing it against other baselines that were recently develop in the learning to learn community to optimize deep neural networks.}

\bibliography{rover_descent_bib}
\bibliographystyle{apalike}

\newpage
\section{Appendix}
{
	\subsection{Landscape generation in details}
	{
		\label{sec::landscape_detail}
		
		We generate random instances of five landscapes: saddles, valleys, plateaus, cliffs and quadratic bowls. The quadratic bowls are generated by sampling random matrices $A$ and vectors $b$ and aggregating them in a quadratic loss $\lVert Ax - b \rVert_2^2$. The other four landscapes are generated by creating random gaussian fields with Mahalanobis-norm covariance functions. More precisely, we carefully sample a given number of points $x$ which are attributed a random value $f(x)$. We then sample a covariance function $k(x,x') = \frac{1}{2} x^T S^{-1} x$ with $S$ being a randomly generated definite positive matrix, carefuly set to generate the targeted landscape.We therefore create a class of functions $\mathcal{F}_{train}$, from which we can sample random instances of the different targeted landscapes. We also add randomness inside each of these instances by not always taking the mean of the generated random fields, but by sampling inside the resulting distribution over space.
		
		We hereinafter describe precisely how we generated samples of each modality in $\mathcal{F}_{train}$ (except quadratic bowls). This procedure is extremely similar to the one that is followed when one makes inference with Gaussian Processes - see \cite{rasmussen2006gaussian}. It starts by carefully sampling a collections of points $\mathcal{X} = \{x_1,\hdots,x_k\}$ and their associated values $\mathcal{V} = \{v_1,\hdots,v_k\}$ to create a specificly targeted landscape. We then sample a positive definite scaling matrix $S$ for the normalized Gaussian covariance function: 
		\begin{equation}
			k(x,x') = \frac{1}{\vert 2\pi S\vert^{1/2}}\exp{\left(-\frac{1}{2}x^TS^{-1}x'\right)}
		\end{equation} 
		Once these steps are completed, and for $F(x) \triangleq \left(f(x),v_1,\hdots,v_k\right)$, we make a Gaussian hypothesis over the joint distribution:
		\begin{equation}
			 p(F) = \mathcal{N}\left( F\, \vert\, 0 , \begin{pmatrix} K \\ k(x) \end{pmatrix} \right)
		\end{equation} 
		where $K = (k(x_i,x_j))_{i,j}$ and $k(x) = (k(x,x_i))_i$. 
		We then can evaluate the conditional distribution $p(f(x) \, \vert \, v_1,\hdots,v_k)$ (which is also a Gaussian) and sample from it to create a noisy version of $f(x)$. We repeat this procedure whenever we need to access to the value of one of the loss in $\mathcal{F}_{train}$ at a point $x$. 
		
		\paragraph{} The following lists details how $\mathcal{X}$ and $\mathcal{V}$ were sampled for each modalities of $\mathcal{F}_{train}$.
		\begin{itemize}
			\item \textbf{Valleys}: We set $\mathcal{X} = \{ 0_{\mathbb{R}^2} \}$ and sample $v_1$ uniformly at random in $[-5,0]$. We sample one value $\lambda_1$ in a positive truncated Gaussian distribution centered at 10 with variance 2. We then multiply it by a ratio $\rho$ uniformly sampled at random inside the interval $[100,200]$ to obtain $\lambda_2 = \rho \lambda_1$. We sample uniformly at random $\phi$ in $[0,2\pi]$, and create $S = R_\phi^T \text{diag}(\lambda_1,\lambda_2) R_\phi$ where $R_\phi$ is the two-dimensional rotation matrix of angle $\phi$. By that mean, we are able to create valleys of different width and orientation.
			\item \textbf{Saddles} : we sample $4$ points $x_1,\hdots, x_4$ uniformly at random within each quarter of the square $[-1,1]^2$ and place them in $\mathcal{X}$. We assign a random value to each one (sampled from a Gaussian distribution) so that to opposite points have values of similar signs. We then sample $\lambda_1,\lambda_2$ in a truncated positive Gaussian distribution centered at $10$ and with variance $2$. We sample a random angle $\phi$ and compute $S = R_\phi^T \text{diag}(\lambda_1,\lambda_2) R_\phi$.
			\item \textbf{Plateau+cliffs}:  $\mathcal{X} = \{ 0_{\mathbb{R}^2} \}$  and generate a single value $v_1$ from a positive truncated Gaussian distribution centered in -5 with variance $2$. We then create a matrix $S$ in the same fashion as for the previously described landscape. 
		\end{itemize}

	}
	
	\subsection{Policy evaluation}
	{
		\label{sec::policy_viz}
		We describe here the learnt policy behavior's for the step-size and resolution updates. The procedure we follow is simple: we sample a functions from each modality of $\mathcal{F}_{train}$ as well as an initial iterate for each one. We then run an iteration procedure and record the evolution of the step-size and the resolution. Figure \ref{fig::policy_seq} display those evolutions  for quadratic bowls, valleys, saddles and (plateaus+cliffs). We can see a variety of behaviors, that tends to remain constant for every instance of a modality. In valleys - \ref{fig::policy_seq_valley}, the incentive of the policy is to keep a large resolution until it detects the optimum is near, in order to get the best representation of the axis of the valley. It then decreases the resolution and the step-size to safely approach the optimum. We see a similar behavior in saddles - \ref{fig::policy_seq_saddle}. For quadratic bowls \ref{fig::policy_seq_quadratic} and (plateaus+cliffs) \ref{fig::policy_seq_cliff}, we see a constant decrease of the resolution in order to be as precise as possible for the angle prediction, and after a acceleration of the step-size we see its global decrease to approach the minimum.  
		
		\begin{figure}[h!]
			\centering
			\begin{subfigure}[b]{0.5\linewidth}
			{
				\centering
				\includegraphics[width=0.8\textwidth]{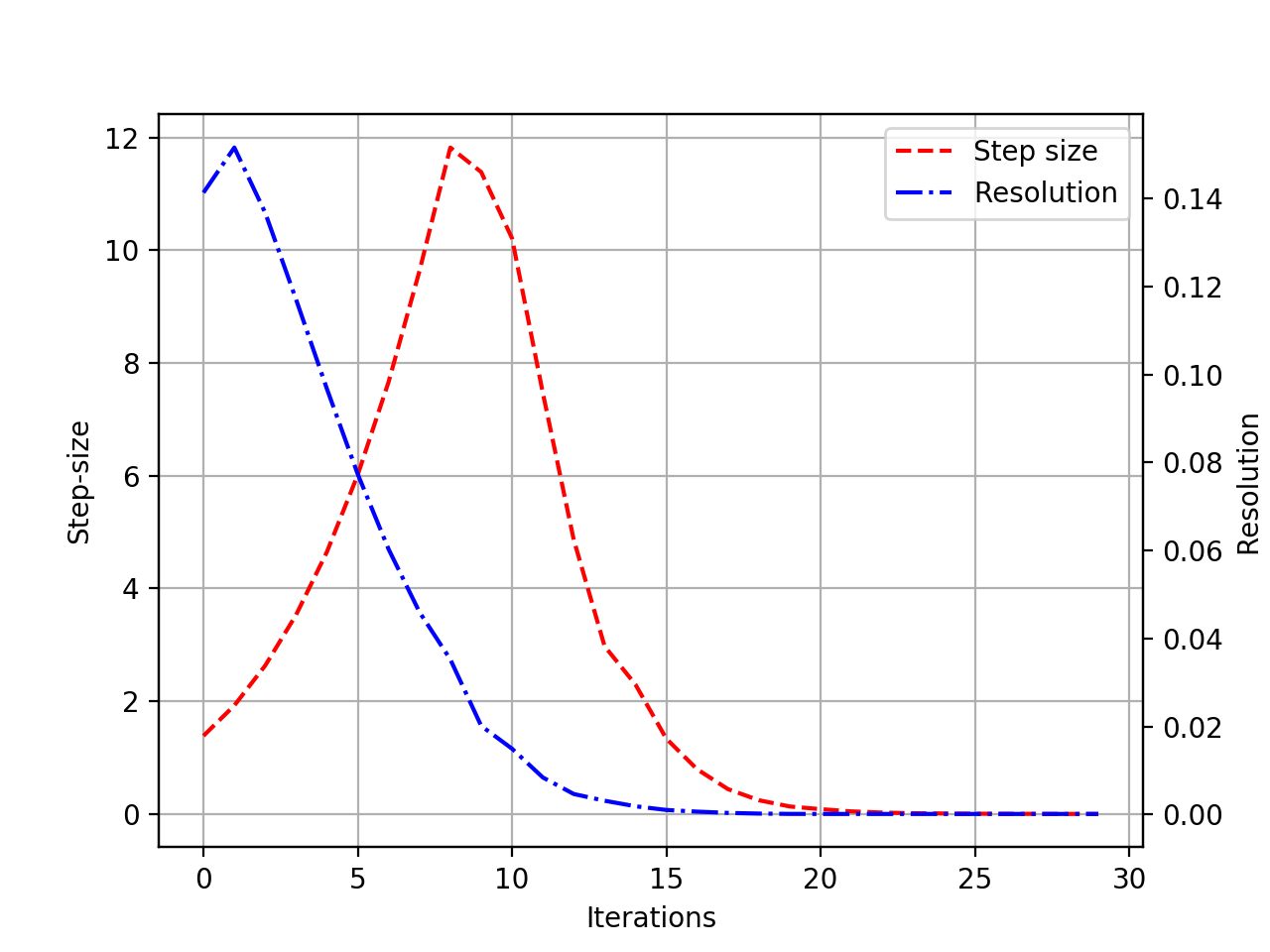}
				\caption{Quadratic}
				\label{fig::policy_seq_quadratic}
			}
			\end{subfigure}\hfill
			\begin{subfigure}[b]{0.5\linewidth}
			{
				\centering
				\includegraphics[width=0.8\textwidth]{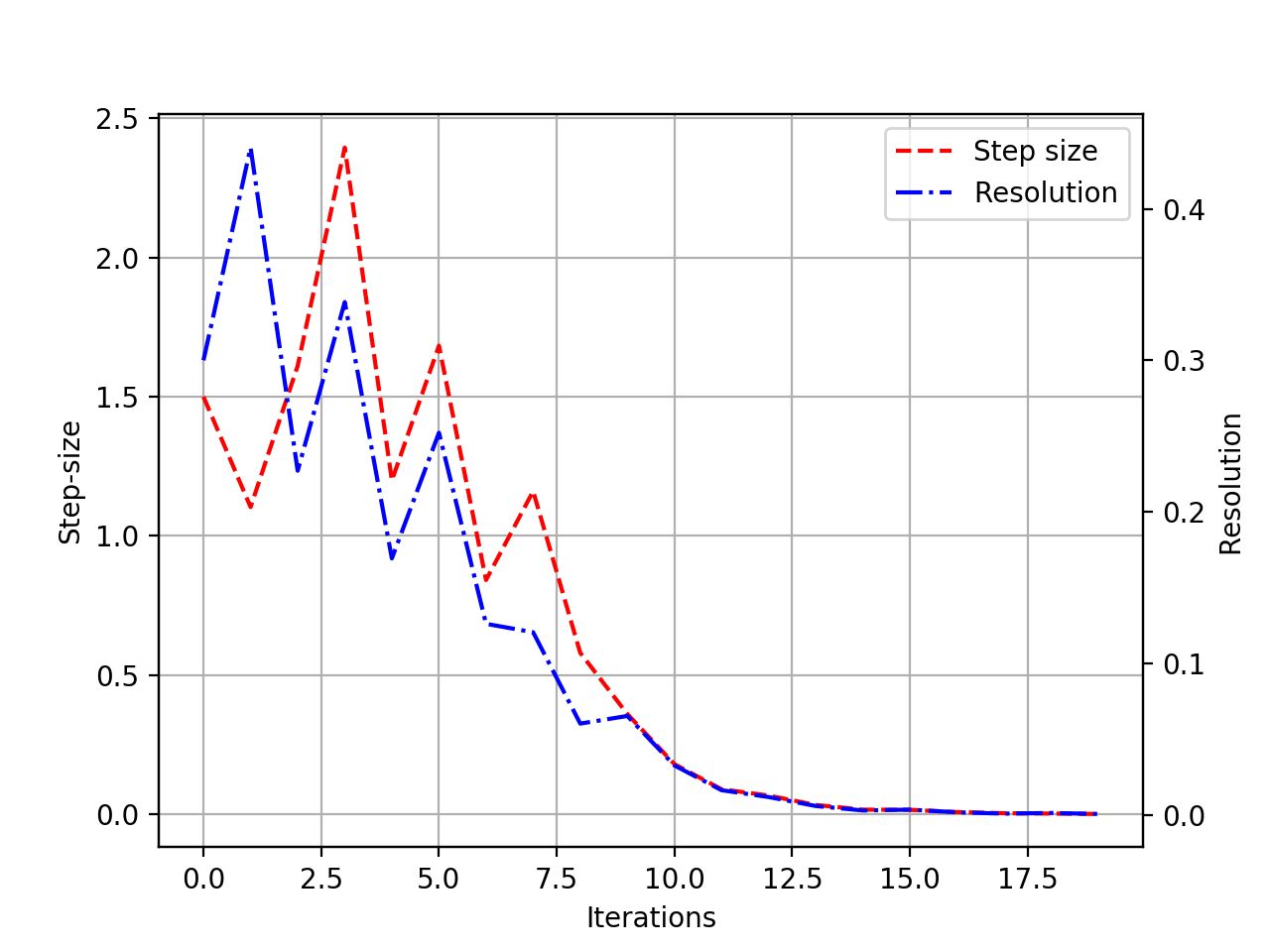}
				\caption{Valley}
				\label{fig::policy_seq_valley}
			}
			\end{subfigure}\\
			\begin{subfigure}[b]{0.5\linewidth}
			{
				\centering
				\includegraphics[width=0.8\textwidth]{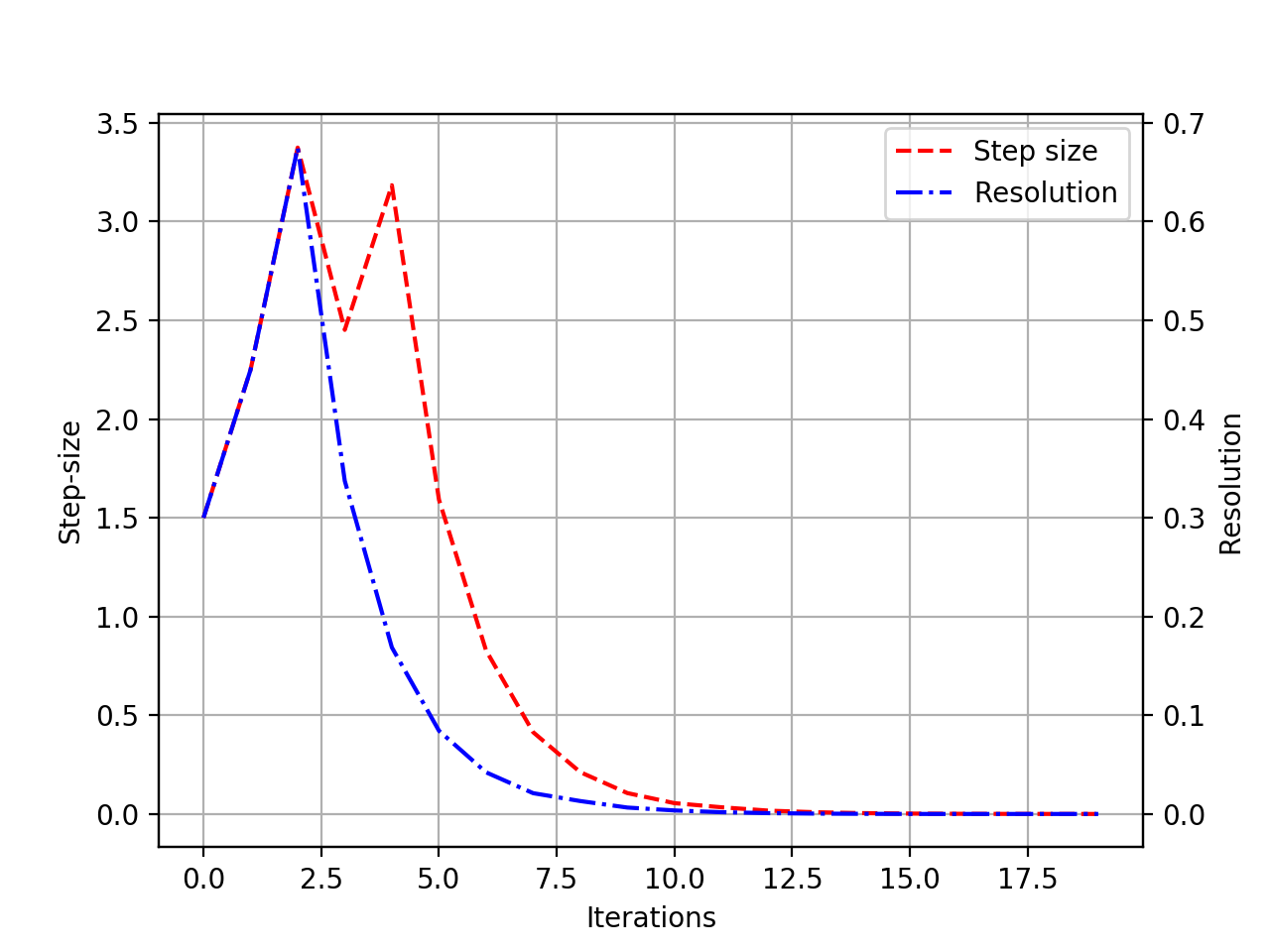}
				\caption{Saddle}
				\label{fig::policy_seq_saddle}
			}
			\end{subfigure}\hfill
			\begin{subfigure}[b]{0.5\linewidth}
			{
				\centering
				\includegraphics[width=0.8\textwidth]{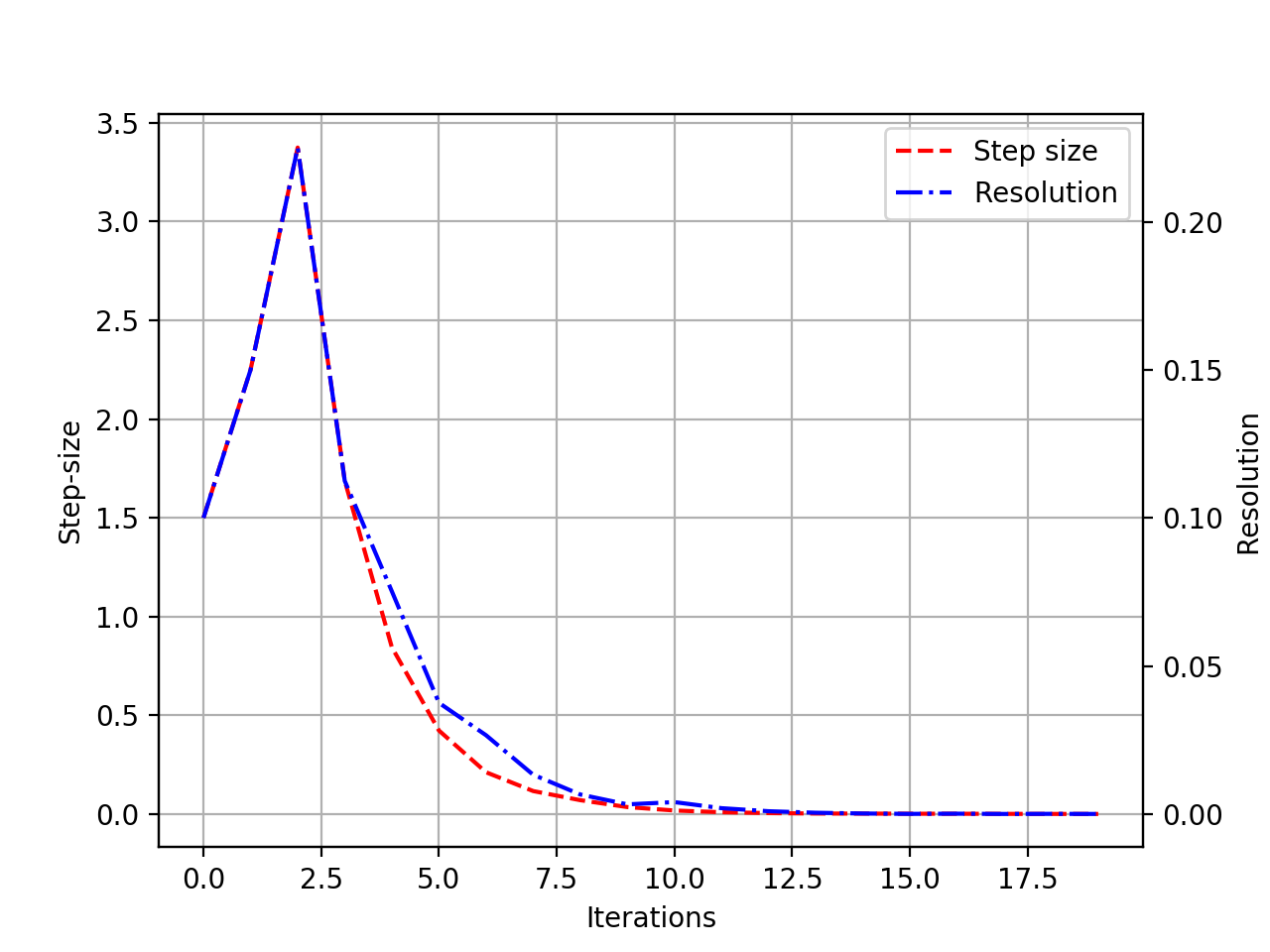}
				\caption{Plateau+cliff}
				\label{fig::policy_seq_cliff}
			}
			\end{subfigure}
			\caption{Step-size and resolution trajectories for different instances of $\mathcal{F}_{train}$}
			\label{fig::policy_seq}
		\end{figure}

		This curves could suggest that the policy has only learnt good scheduling schemes for each of the modalities in $\mathcal{F}_{train}$. We show here that this is not the case, by showing that the policy is able to adapt to sharp changes of its initial state. In Figure \ref{fig::policy_seq_ad}, we sample this initial state far from the distribution that was used at training time (typically, a learning rate that is 100 times smaller that it usually is at training time, and a resolution 100 bigger), for a quadratic bowl. We can see that the schedule adapts to these changes, and therefore that only changes in the local landscape seen by the agent affects its decision (and not a fixed schedule learnt by the RNN). 
		
		\begin{figure}[h!]
			\centering
			\includegraphics[width=0.4\linewidth]{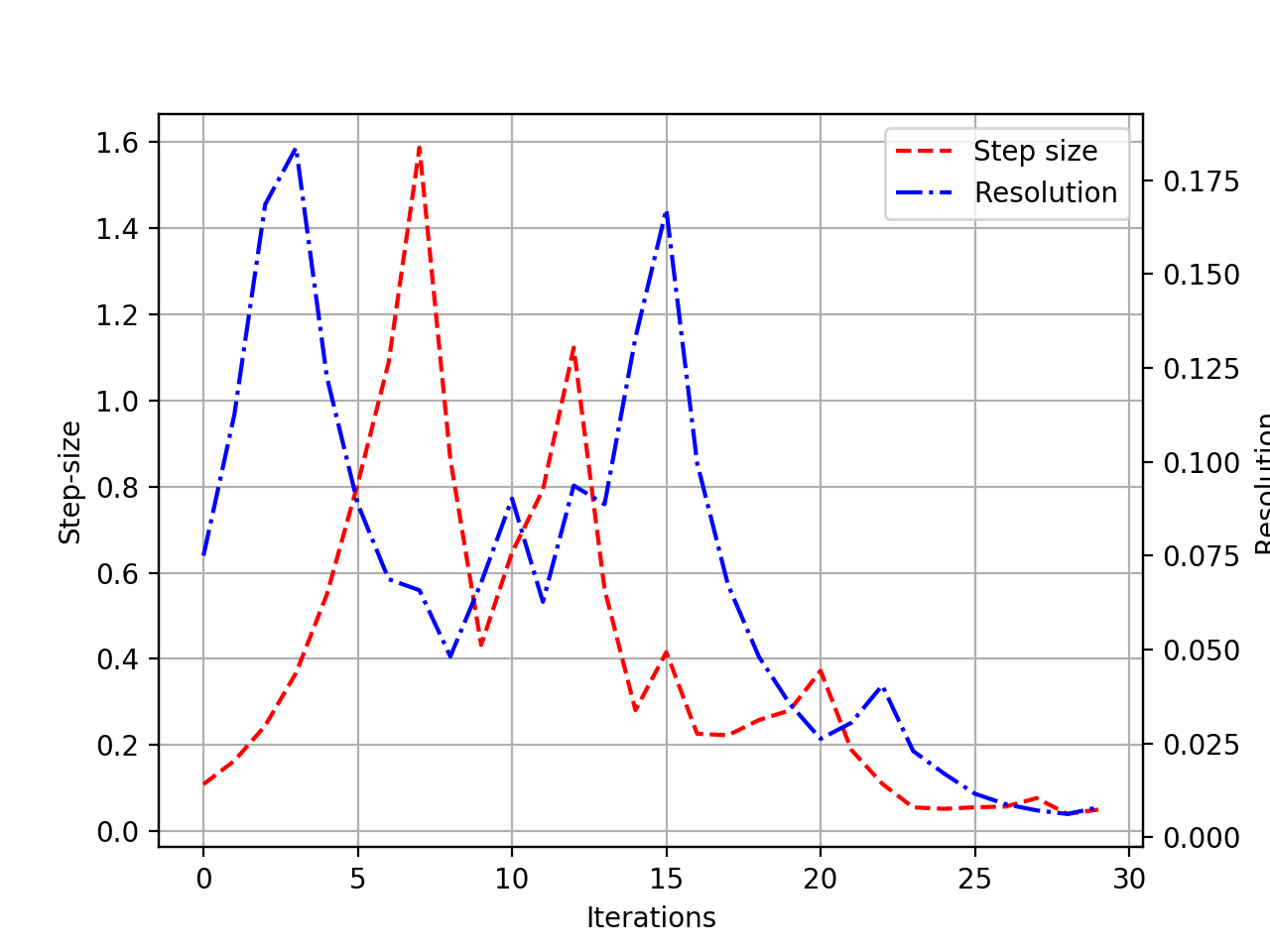}
			\caption{Step-size and resolution can recover from a poorly set initial state}
			\label{fig::policy_seq_ad}
		\end{figure}	

		\paragraph{} We also show in Figure \ref{fig::policy_seq_contour} similar runs on contour plots, with a visualization of the sample grid. 
		\begin{figure}[h!]
			\centering
			\begin{subfigure}[b]{0.4\linewidth}
			{
				\centering
				\includegraphics[width=0.8\textwidth]{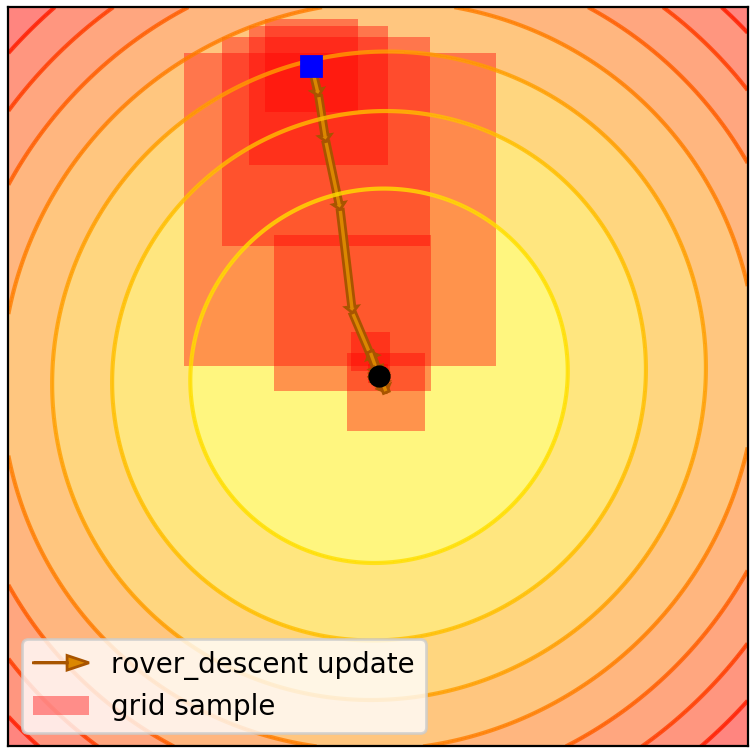}
				\caption{Quadratic}
			}
			\end{subfigure}
			\begin{subfigure}[b]{0.4\linewidth}
			{
				\centering
				\includegraphics[width=0.8\textwidth]{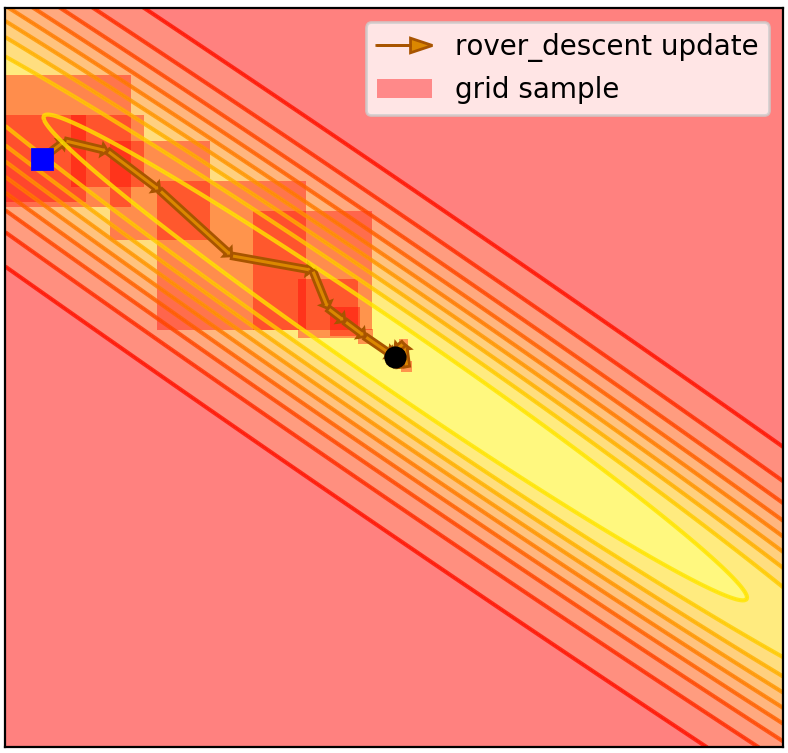}
				\caption{Valley}
			}
			\end{subfigure}\\
			\begin{subfigure}[b]{0.4\linewidth}
			{
				\centering
				\includegraphics[width=0.8\textwidth]{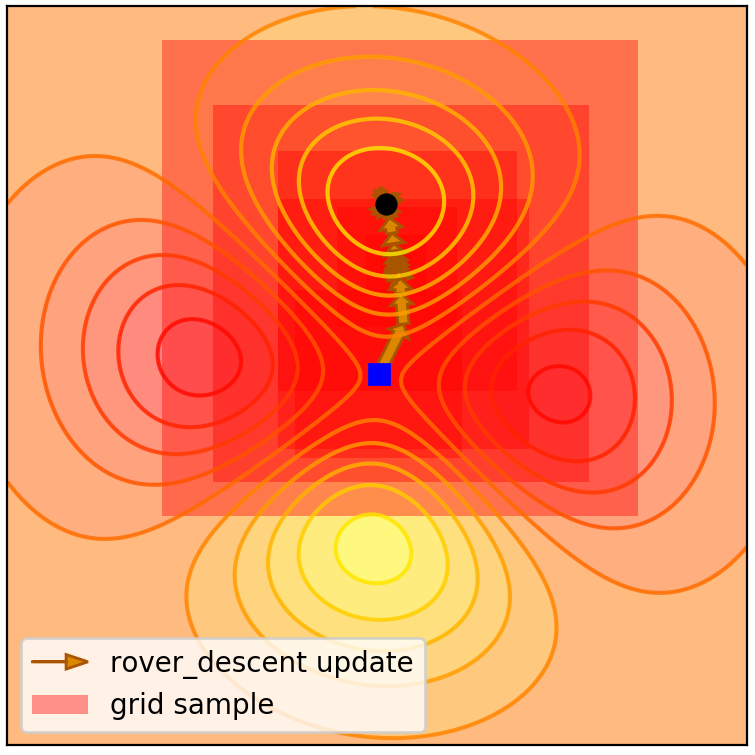}
				\caption{Saddle}
				\label{fig::policy_seq_saddle}
			}
			\end{subfigure}
			\begin{subfigure}[b]{0.4\linewidth}
			{
				\centering
				\includegraphics[width=0.8\textwidth]{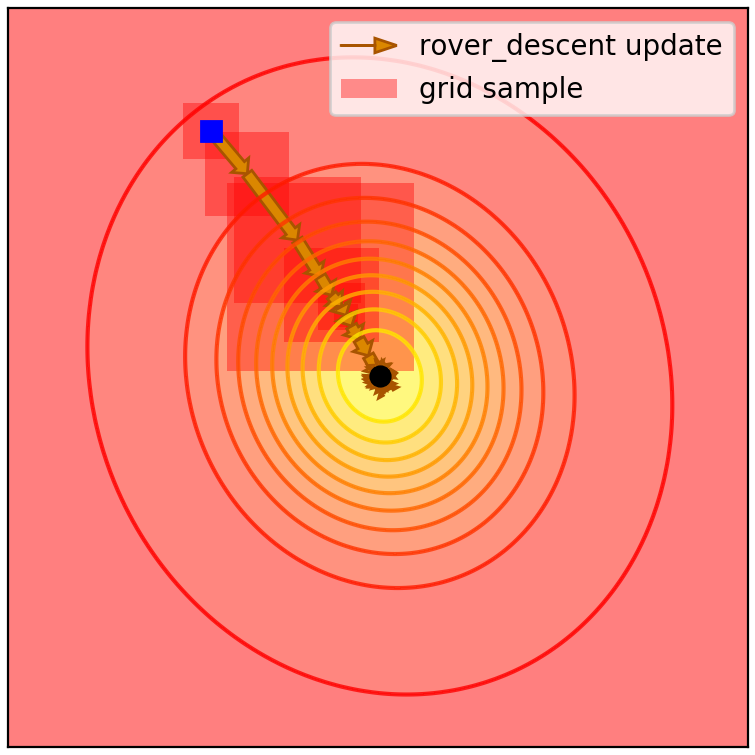}
				\caption{Plateau+cliff}
				\label{fig::policy_seq_cliff}
			}
			\end{subfigure}
			\caption{Contour plot of optimization runs for policy vizualisation on $\mathcal{F}_{train}$}
			\label{fig::policy_seq_contour}
		\end{figure}
		
	}

	\subsection{Precisions on $\mathcal{F}_{test}$}
	{
		\label{sec::ftest}
		We provide here some contour plots for the functions used in the meta-test dataset. We also provide their mathematical expression, as well as the position of their global optimum and the position of the initial iterate we used in our experiment. 
		
		\subsubsection{Rosenbrock's function}
		{
			Rosenbrock's function's analytical expression is:
			\begin{equation}
				f(x) = 100(x_1-x_0^2)^2 + (x_0-1)^2
			\end{equation}
			and its contour plot is shown in \ref{fig::rosenbrock}.
			
			\begin{figure}[h!]
				\begin{center}
					\includegraphics[width=0.3\linewidth]{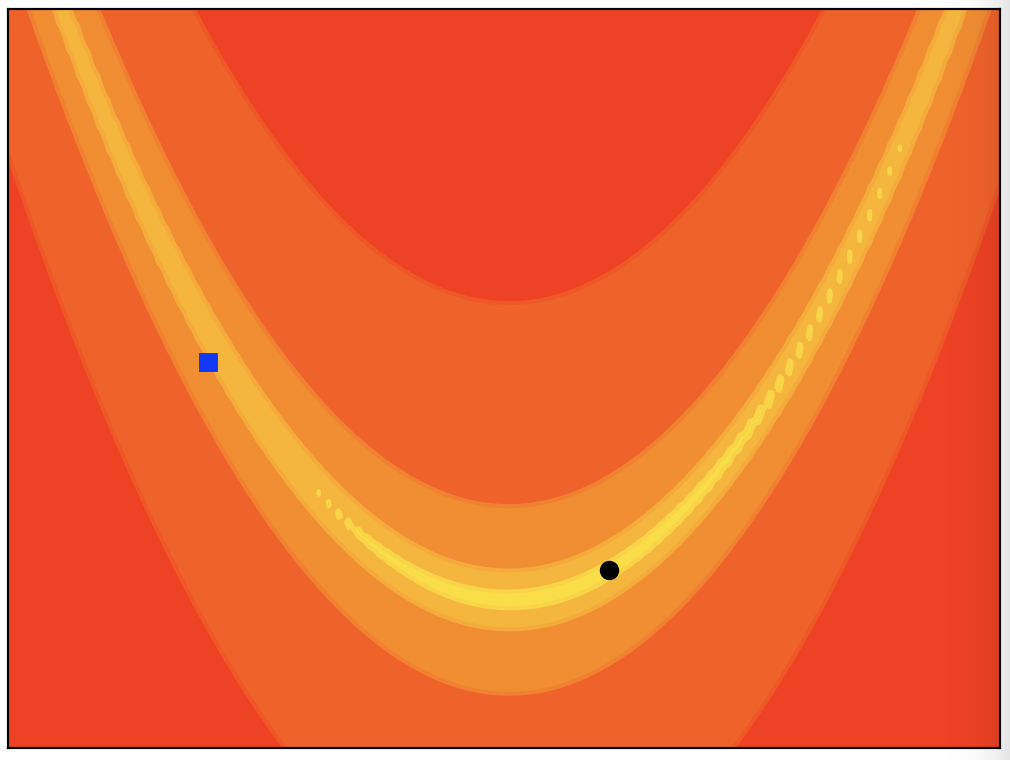}
					\caption{Contour plot of the Rosenbrock's function. The black circle indicates the position of the optimum and the blue square the position of the initial iterate.}
					\label{fig::rosenbrock}
				\end{center}
			\end{figure}
			
		}
		
		\subsubsection{Ackley's function}
		{
			Acley's function's analytical expression is:
			\begin{equation}
				f(x) = -20\exp{\left(-0.2\sqrt{\frac{1}{2}(x_1^2+x_2^2)}\right)} - \exp{\left(\frac{1}{2}\cos{2\pi x_1} + \frac{1}{2}\cos{2\pi x_2}\right)} + 20 + e^1
			\end{equation}
			and its contour plot is shown in \ref{fig::ackley}.
			
			\begin{figure}[h!]
				\begin{center}
					\includegraphics[width=0.3\linewidth]{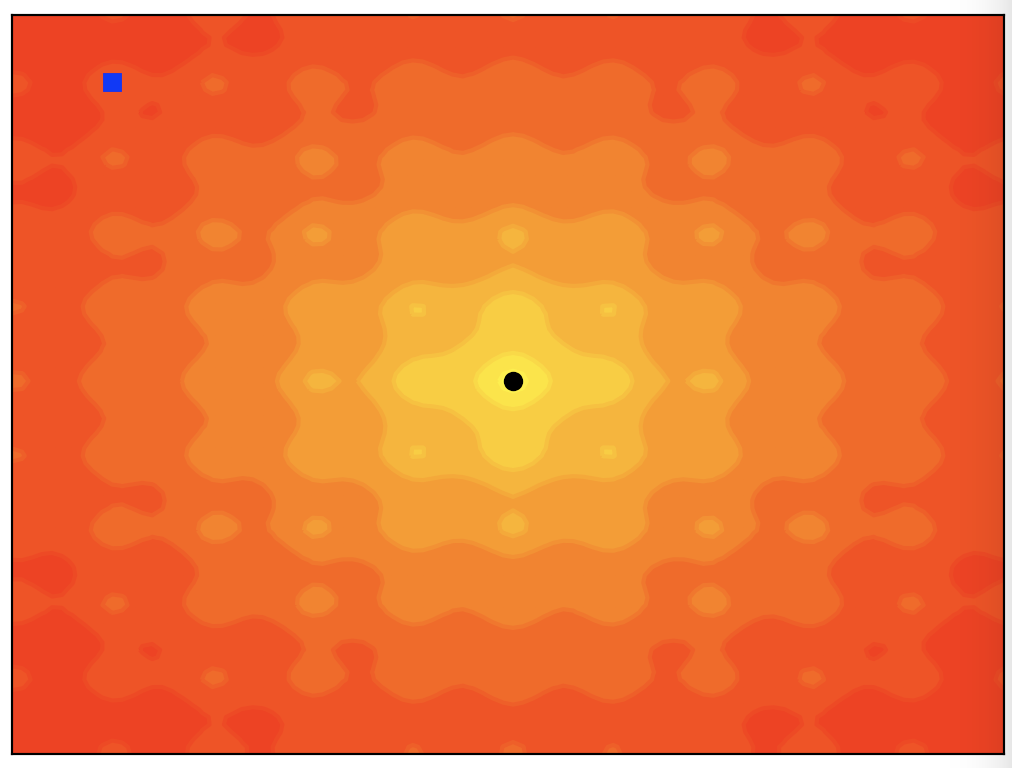}
					\caption{Contour plot of the Ackley's function. The black circle indicates the position of the optimum and the blue square the position of the initial iterate.}
					\label{fig::ackley}
				\end{center}
			\end{figure}
		}
		
		\subsubsection{Rastrigin's function}
		{
			Rastrigin's function's analytical expression is:
			\begin{equation}
				f(x) = 20 + \sum_{i=1}^2 (x_i^2 - 10\cos{(2\pi x_i)})
			\end{equation}
			and its contour plot is shown in \ref{fig::Rastrigin}.
			
			\begin{figure}[h!]
				\begin{center}
					\includegraphics[width=0.3\linewidth]{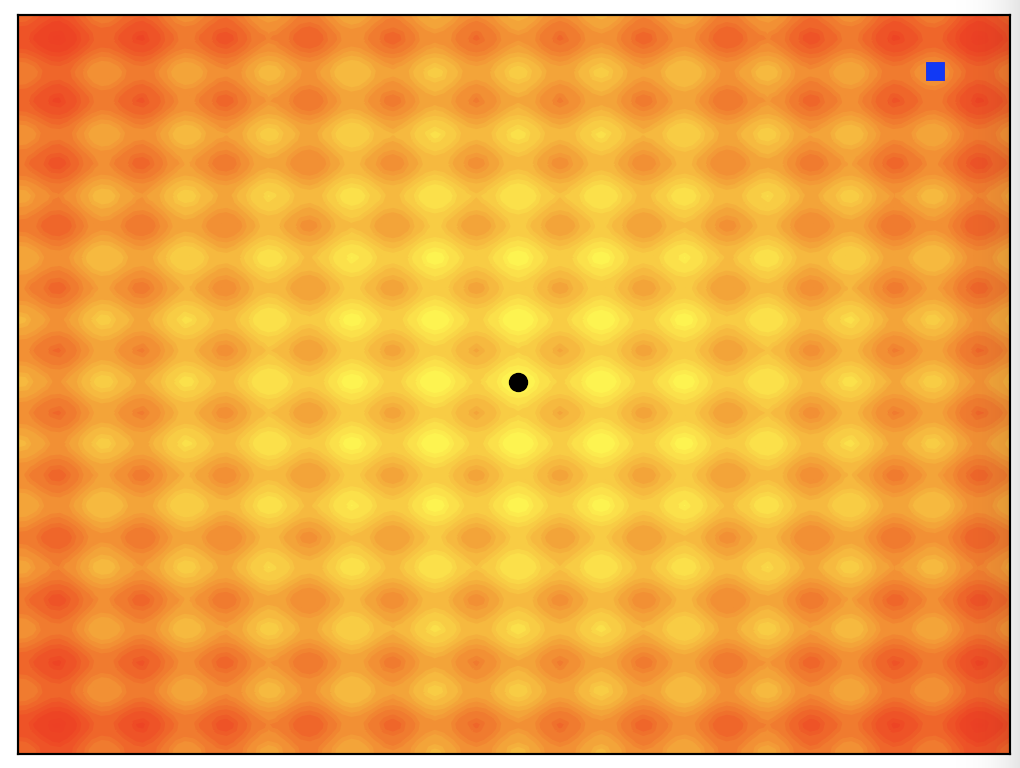}
					\caption{Contour plot of the Rastrigin function. The black circle indicates the position of the optimum and the blue square the position of the initial iterate.}
					\label{fig::Rastrigin}
				\end{center}
			\end{figure}
		}
		
		\subsubsection{Maccornick's function}
		{
			Maccornick's function's analytical expression is:
			\begin{equation}
				f(x) = \sin{(x_1+x_2)} + (x_1-x_2)^2 -1.5x_1 + 2.5x_2 +1 
			\end{equation}
			and its contour plot is shown in \ref{fig::Maccornick}.
			
			\begin{figure}[h!]
				\begin{center}
					\includegraphics[width=0.3\linewidth]{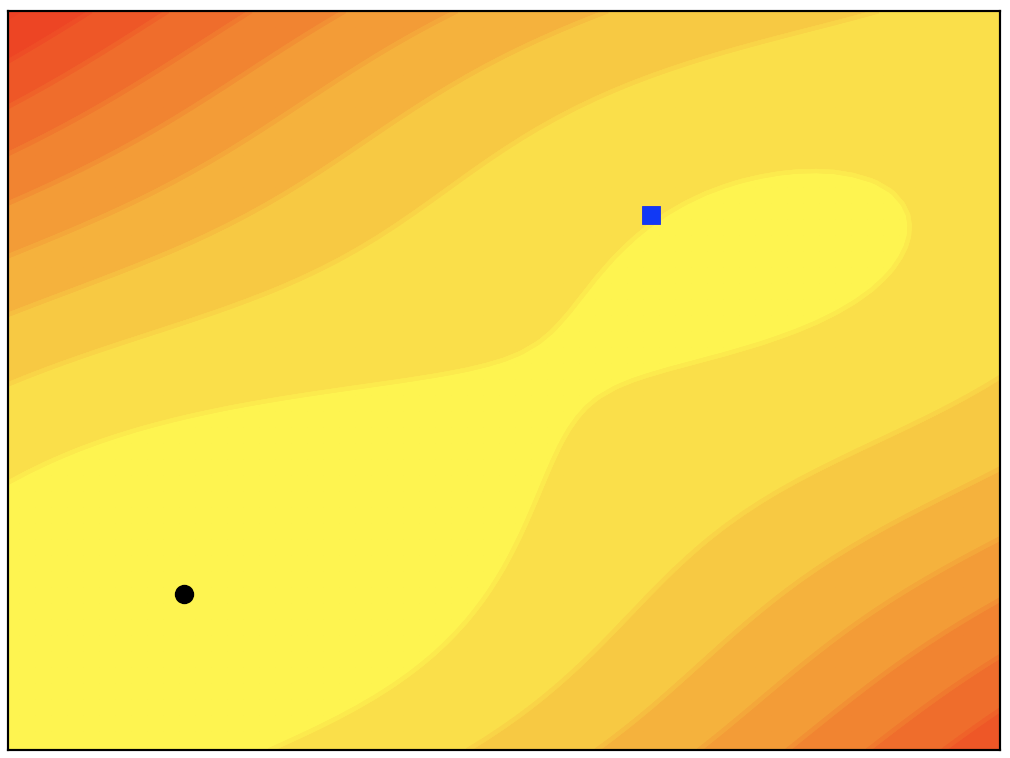}
					\caption{Contour plot of the Maccornick function. The black circle indicates the position of the optimum and the blue square the position of the initial iterate.}
					\label{fig::Maccornick}
				\end{center}
			\end{figure}
		}
		
		\subsubsection{Styblinski's function}
		{
			Styblinski's function's analytical expression is:
			\begin{equation}
				f(x) = \frac{1}{2} \sum_{i=1}^2 (x_i^4 - 16x_i^2+5x_i)
			\end{equation}
			and its contour plot is shown in \ref{fig::Styblinski}.
			
			\begin{figure}[h!]
				\begin{center}
					\includegraphics[width=0.3\linewidth]{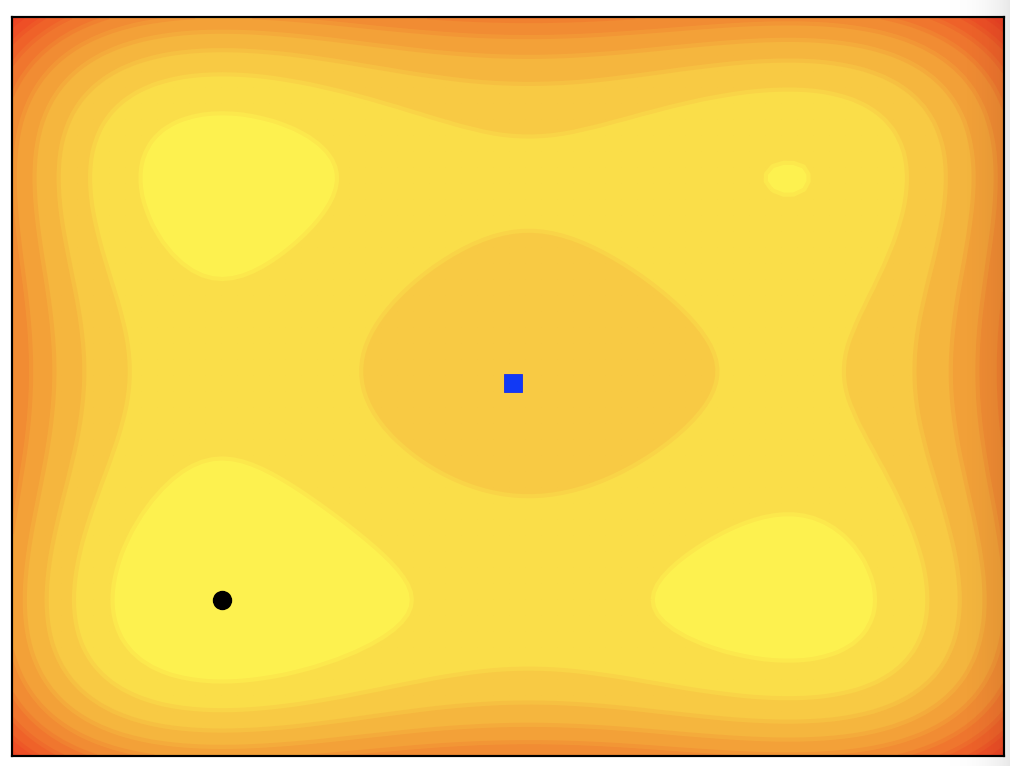}
					\caption{Contour plot of the Styblinski function. The black circle indicates the position of the optimum and the blue square the position of the initial iterate.}
					\label{fig::Styblinski}
				\end{center}
			\end{figure}
		}
		
		\subsubsection{Beale's function}
		{
			Beale's function's analytical expression is:
			\begin{equation}
				f(x) = (1.5-x_1+x_1x_2)^2 + (2.25-x_1+x_1x_2^2)^2 + (2.625-x_1+x_1x_2^3)^2
			\end{equation}
			and its contour plot is shown in \ref{fig::Beale}.
			
			\begin{figure}[h!]
				\begin{center}
					\includegraphics[width=0.3\linewidth]{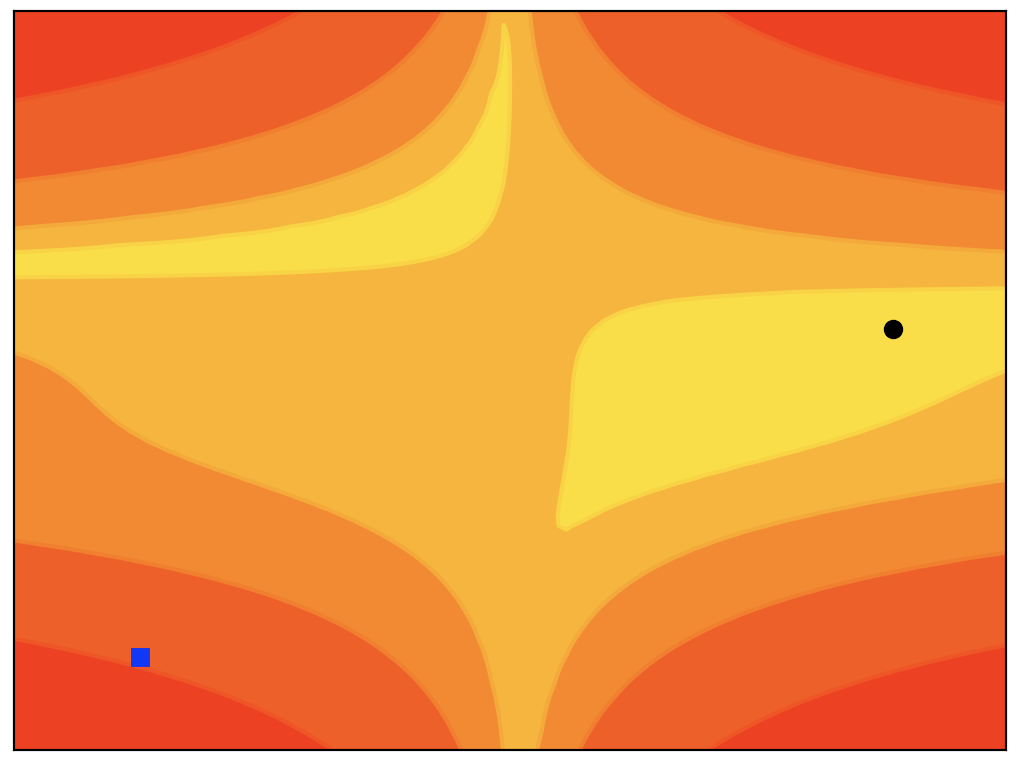}
					\caption{Contour plot of the Beale function. The black circle indicates the position of the optimum and the blue square the position of the initial iterate.}
					\label{fig::Beale}
				\end{center}
			\end{figure}
		}
	}
}

\end{document}